\RequirePackage{iftex}
\ifPDFTeX
  \pdfoutput=1
\fi
\documentclass[10pt,logo,copyright]{nvidiatechreport}
\usepackage[numbers,sort&compress,square]{natbib}

\ifPDFTeX
  \usepackage[utf8]{inputenc} %
  \usepackage[T1]{fontenc}    %
\fi

\usepackage{parskip}        %
\usepackage{url}            %
\usepackage{hyperref}       %
\usepackage{booktabs}       %
\usepackage{amsfonts}       %
\usepackage{nicefrac}       %
\usepackage{microtype}      %
\usepackage{xcolor}         %
\usepackage[dvipsnames]{xcolor} %
\usepackage{graphicx}
\usepackage{subcaption}
\usepackage{tabularx}
\usepackage{makecell}
\usepackage{adjustbox}
\usepackage{setspace}
\newcolumntype{M}[1]{>{\centering\arraybackslash}m{#1}}
\usepackage{float}
\usepackage{amsmath,amsfonts,bm}
\IfFileExists{bbm.sty}{\usepackage{bbm}}{}
\usepackage{multirow}
\usepackage{comment}
\usepackage{lipsum}
\usepackage{wrapfig}
\usepackage{caption}
\IfFileExists{siunitx.sty}{\usepackage{siunitx}}{}
\usepackage{xspace}
\IfFileExists{algorithm.sty}{\usepackage{algorithm}}{
\usepackage{float}
\floatstyle{ruled}
\newfloat{algorithm}{thp}{loa}
\floatname{algorithm}{Algorithm}

\newenvironment{algorithmic}[1][]{\par\begin{quote}\ttfamily\small}{\end{quote}\par}
\providecommand{\State}{\par}
\providecommand{\Procedure}[2]{\textbf{procedure} \textsc{#1}#2\par}
\providecommand{\EndProcedure}{\textbf{end procedure}\par}
\providecommand{\If}[1]{\textbf{if}\ #1\ \textbf{then}\par}

\providecommand{\EndIf}{\textbf{end if}\par}
\providecommand{\For}[1]{}\renewcommand{\For}[1]{\textbf{for}\ #1\ \textbf{do}\par}
\providecommand{\EndFor}{\textbf{end for}\par}
\providecommand{\While}[1]{\textbf{while}\ #1\ \textbf{do}\par}
\providecommand{\EndWhile}{\textbf{end while}\par}
\providecommand{\Return}{\textbf{return}\ }
\providecommand{\Comment}[1]{\hfill\textit{#1}}
}
\IfFileExists{algpseudocode.sty}{\usepackage[noend]{algpseudocode}}{}

\definecolor{darkred}{rgb}{0.7, 0.0, 0.0}

\usepackage{pifont}
\newcommand{\cmark}{\ding{51}} %

\makeatletter
\@ifpackageloaded{siunitx}{%
  \sisetup{
    separate-uncertainty = true,
    table-number-alignment = center,
  }%
}{}
\makeatother

\ifPDFTeX
\else
  \hypersetup{
    pdfauthor={HumanoidMimicGen Authors},
    pdftitle={HumanoidMimicGen: Data Generation for Loco-Manipulation via Whole-Body Planning},
    pdfsubject={Robotics, humanoid loco-manipulation, data generation},
    pdfkeywords={humanoid; loco-manipulation; data generation; imitation learning},
  }
\fi

\newcommand{\sysName}{HumanoidMimicGen\xspace}

\newcommand{\proc}[1]{\textsc{#1}}
\newcommand{\kw}[1]{{\bf #1}}

\newcommand{\R}{\mathbf{R}}

\newcommand{\SE}[1]{\mathrm{SE}(#1)}

\newcommand{\inv}[1]{{#1}^{-1}}
\DeclareMathOperator\supp{supp}

\newif\ifcomments
\commentsfalse

\ifcomments
    \newcommand{\todo}[1]{\textcolor{red}{(TODO: #1)}}
    \newcommand{\caelan}[1]{\textcolor{blue}{(CG: #1)}}
    \newcommand{\ajay}[1]{\textcolor{green}{(AM: #1)}}
    \newcommand{\kevin}[1]{\textcolor{cyan}{(KL: #1)}}
    \newcommand{\yuke}[1]{\textcolor{magenta}{(YZ: #1)}}
\else
    \newcommand{\todo}[1]{}
    \newcommand{\caelan}[1]{}
    \newcommand{\ajay}[1]{}
    \newcommand{\kevin}[1]{}
    \newcommand{\yuke}[1]{}
\fi

\newcommand{\projectlink}{https://humanoidmimicgen.github.io/}

\title{HumanoidMimicGen: Data Generation for Loco-Manipulation via Whole-Body Planning}

\author{
    Kevin Lin$^{1,2,\ast}$, Ajay Mandlekar$^{1,\ast}$, Caelan Reed Garrett$^{1,\ast}$,
    Nikita Chernyadev$^{1}$, Yu Fang$^{1}$, Runyu Ding$^{1}$,
    Yuqi Xie$^{1}$, Justin Tran$^{1}$, Linxi Fan$^{1,\dagger}$, Yuke Zhu$^{1,2,\dagger}$ \\
    \small $^{1}$NVIDIA \quad $^{2}$The University of Texas at Austin \\
    \small $^{\ast}$ Equal contribution \quad $^{\dagger}$ Project leads \\
    \small \tt{\href{\projectlink}{\projectlink}}
}

\begin{document}
\maketitle

\begin{abstract}
Imitation learning is a promising approach for training humanoid robots to both walk and manipulate, but it requires a large number of demonstrations, which are time-intensive and difficult to collect via teleoperation. 
Existing data-generation algorithms can automatically synthesize demonstrations for manipulators, but they are ineffective on humanoids because their high-dimensional composite action spaces involve arms, legs, and torsos.
We present \sysName{}, a method for generating humanoid legged loco-manipulation data.
Our method adapts contact-rich whole-body skills from a handful of source demonstrations to new states, generalizing across changes in object pose.
By interleaving these single- and dual-arm skills with whole-body locomotion and manipulation planning, the method generates stable, collision-free data across diverse scenes and layouts.
To evaluate our approach, we introduce a new simulated loco-manipulation benchmark containing nine diverse tasks that test humanoid loco-manipulation capabilities. 
There, we demonstrate that \sysName{} automatically generates large datasets for imitation learning and enables a systematic study of how data generation and policy learning decisions impact model performance. 
We show that whole-body visuomotor policies co-trained with data generated by \sysName{} outperform those trained only on real-world data by \textbf{20\%}. 
\end{abstract}
\abscontent

\section{Introduction}

Vision-Language-Action (VLA) models~\cite{bjorck2025gr00t,black2024pi_0}, trained on large robotic manipulation datasets, have recently demonstrated a remarkable capability to enable robots to autonomously complete a wide range of manipulation tasks. 
However, this paradigm has largely been demonstrated in stationary robot manipulation settings, where the robot does not need to walk around in its workspace. 
This contrasts with the promise of humanoid robots: their human-like form factor should enable them to move and interact fluidly with environments designed for humans. 
However, applying the VLA paradigm to humanoid loco-manipulation remains challenging due to two key difficulties.
First, VLAs require large-scale manipulation datasets to develop their capabilities. Sourcing these datasets for even stationary manipulation settings remains a challenging endeavor due to the persistent human time and effort required.
The data collection process typically involves robot teleoperation, where teams of human operators control fleets of robots to collect datasets over months~\cite{ebert2021bridge,brohan2022rt,o2024open,khazatsky2024droid}. 
Second, intelligently controlling a humanoid is a difficult problem due to its high degrees of freedom and the need to balance itself during simultaneous manipulation and locomotion.
Consequently, large-scale loco-manipulation datasets are not readily available, making VLA training prohibitive.

Simulation and synthetic data generation offer a compelling means to address the challenge of sourcing large-scale datasets for training VLAs. 
Automated data generation tools~\cite{mandlekar2023mimicgen, dalal2023imitating} allow for generating large volumes of data, and furthermore, recent results have shown that these synthetic datasets can train real-world manipulation policies~\cite{maddukuri2025sim, wei2025empirical, cheng2025generalizable, haldar2026point}. 
Furthermore, such tools can enable studies into how data quality and policy learning decisions can impact model performance~\cite{robomimic2021, saxena2025matters}.
This kind of systematic investigation is needed for the loco-manipulation setting due to the lack of available datasets and the difficulty of evaluating VLA models in the real world~\cite{barreiros2025careful}.

\begin{figure}[t]
    \centering
    \includegraphics[width=\textwidth]{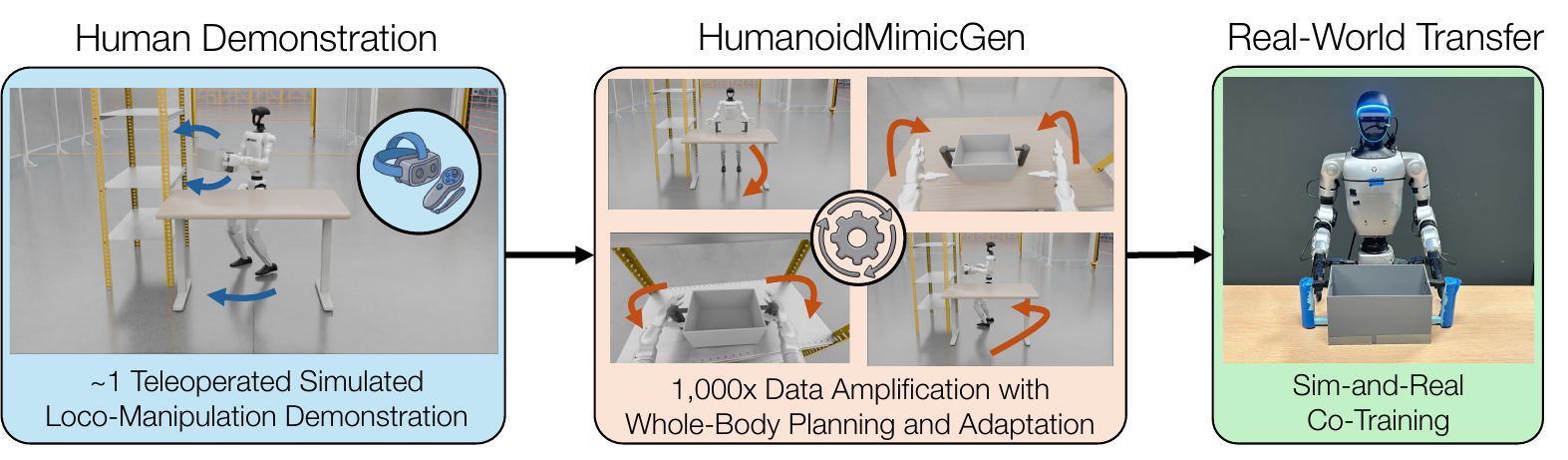}
    \caption{\small \textbf{\sysName{} Overview.} We present \sysName{}, a method for humanoid legged loco-manipulation demonstration generation. \textit{Left}: a human teleoperator collects a loco-manipulation task demonstration. \textit{Center}: \sysName{} generates thousands of demonstrations across scene and object layouts by adapting local segments of the human demonstration via whole-body planning and control. \textit{Right}: these demonstrations train performant loco-manipulation policies in simulation and real-world settings.}
    \label{fig:teaser}
\end{figure}

To address this, we present \sysName{}, \textbf{a method for humanoid legged loco-manipulation demonstration generation.}
\sysName{} interleaves locomotion planning, arm planning, and skill demonstration adaptation to generate loco-manipulation demonstrations in new scenes (see Fig.~\ref{fig:teaser}).
We propose a hybrid control space for humanoids, where legs are controlled by a reinforcement-learned policy and, in turn, adopt a decoupled approach that factors planning into lower- and upper-body motion, corresponding to segments that require dynamic and static stability.
To study data quality and policy training questions in the setting of loco-manipulation, we introduce a new simulation benchmark of humanoid factory tasks.
We show that \sysName{} can generate high-quality data and train performant policies in this setting.
Finally, we demonstrate that generating large simulation datasets with \sysName{} enables sim-and-real co-training for sample-efficient deployment on real-world humanoids.
\textbf{The contributions of this paper are:}

\textbf{\sysName{}:} a data generation method for humanoids that uses whole-body skill planning to synthesize loco-manipulation demonstrations from a few human demonstrations.

\textbf{G1 loco-manipulation benchmark:} a new nine-task simulation benchmark suite of G1 humanoid loco-manipulation tasks, enabling reproducible comparison across methods.

\textbf{Humanoid policy learning analysis:} experiments on this benchmark that study the effect of randomization types, data-generation algorithms, and model architectures on policy success rates.

\textbf{Sim-and-real co-training:} real-world results demonstrating that policies co-trained on real-world data and data generated by \sysName{} outperform policies trained only on real-world data by $20\%$.

\section{Related Work}

\textbf{Automated Data Generation.}
MimicGen~\cite{mandlekar2023mimicgen} introduced a data generation method for generating large-scale datasets by adapting a small set of expert demonstrations to new object poses. 
This method has been extended to support stationary bimanual dexterous manipulation~\cite{jiang2024dexmimicgen} and incorporate planning and optimization techniques~\cite{garrett2024skillgen, yang2025physics, lin2025cpgen} and Reinforcement Learning (RL)~\cite{zhou2025reinforcegen} to improve data quality. 
MoMaGen~\cite{li2025momagen} applied planning and skill adaptation~\cite{garrett2024skillgen} to mobile manipulation and incorporated soft visibility constraints to aid visuomotor policy learning.
WBCMimicGen~\cite{liu2025manipulation} also addressed mobile manipulation, but 
focused on a quadratic programming approach to improving its tracking controller.
Task and Motion Planning (TAMP)~\cite{garrett2021integrated} algorithms have been used for data generation~\cite{mcdonald2022guided,dalal2023imitating,mandlekar2023hitltamp}; but, they require a planning model, which is prohibitive for contact-rich tasks.
These approaches all rely on the assumption of stable and independent end-effector control per limb, which does not transfer to legged humanoids, where careful whole-body coordination is required during both locomotion and manipulation.

\textbf{Learning Manipulation from Demonstrations.} %
Behavior Cloning (BC)~\cite{pomerleau1989alvinn, schaal1999imitation, Ijspeert2002MovementIW, robomimic2021, chi2023diffusion} is a key method for training policies from demonstrations via supervised learning. 
Though this method has proven to be effective for robot manipulation~\cite{Billard2008RobotPB, Calinon2010LearningAR, black2024pi_0, khazatsky2024droid, brohan2023rt, bjorck2025gr00t}, its success relies on having large-scale high-quality datasets to learn from.
Recent evidence has shown that simulation datasets can supplement or replace real-world human demonstrations to produce high-performing real-world manipulation policies~\cite{maddukuri2025sim, wei2025empirical, bjorck2025gr00t, cheng2025generalizable, tian2025interndata, yin2026genie, haldar2026point}.
Here, we use sim-and-real co-training to train real-world policies.

\textbf{Loco-Manipulation.}

Planning and control for humanoid loco-manipulation predates recent learning systems.
Early work studied manipulating objects while walking and integrating dynamic walking with whole-body motion planning~\cite{dalibard2010manipulation,dalibard2013dynamic}.
Other lines formulated humanoid manipulation as multimodal or whole-body planning, including mode-switching manipulation tasks~\cite{Hauser2011Multi-modalTask,HauserIJRR11}, articulated-object manipulation~\cite{burget2013wholebody}, whole-body constrained manipulation~\cite{grey2016humanoid}, and coupled footstep and whole-body planning~\cite{asif2019wholebody}.
More recent classical systems use graph search and multi-contact planning and control for humanoid loco-manipulation~\cite{murooka2021humanoid,ferrari2023multicontact}.
Recent benchmarks and surveys study whole-body locomotion and manipulation in simulation~\cite{sferrazza2024humanoidbench,luo2024smplolympics,gu2025humanoid}.
We build on recent humanoid loco-manipulation work combining teleoperation and reinforcement learning.
Homie~\cite{ben2025homie} and HOVER~\cite{he2025hover} develop whole-body control frameworks integrating locomotion and manipulation, enabling stable execution of loco-manipulation behaviors under learned or hybrid controllers. 
SONIC~\cite{luo2025sonic} scales control to a diverse range of loco-manipulation behaviors and poses.
OmniRetarget~\cite{yang2025omniretarget} trains loco-manipulation policies using reinforcement learning and deploys them in the real world. We use imitation learning to avoid reward tuning.

\begin{figure} %
\centering
\includegraphics[width=\linewidth]{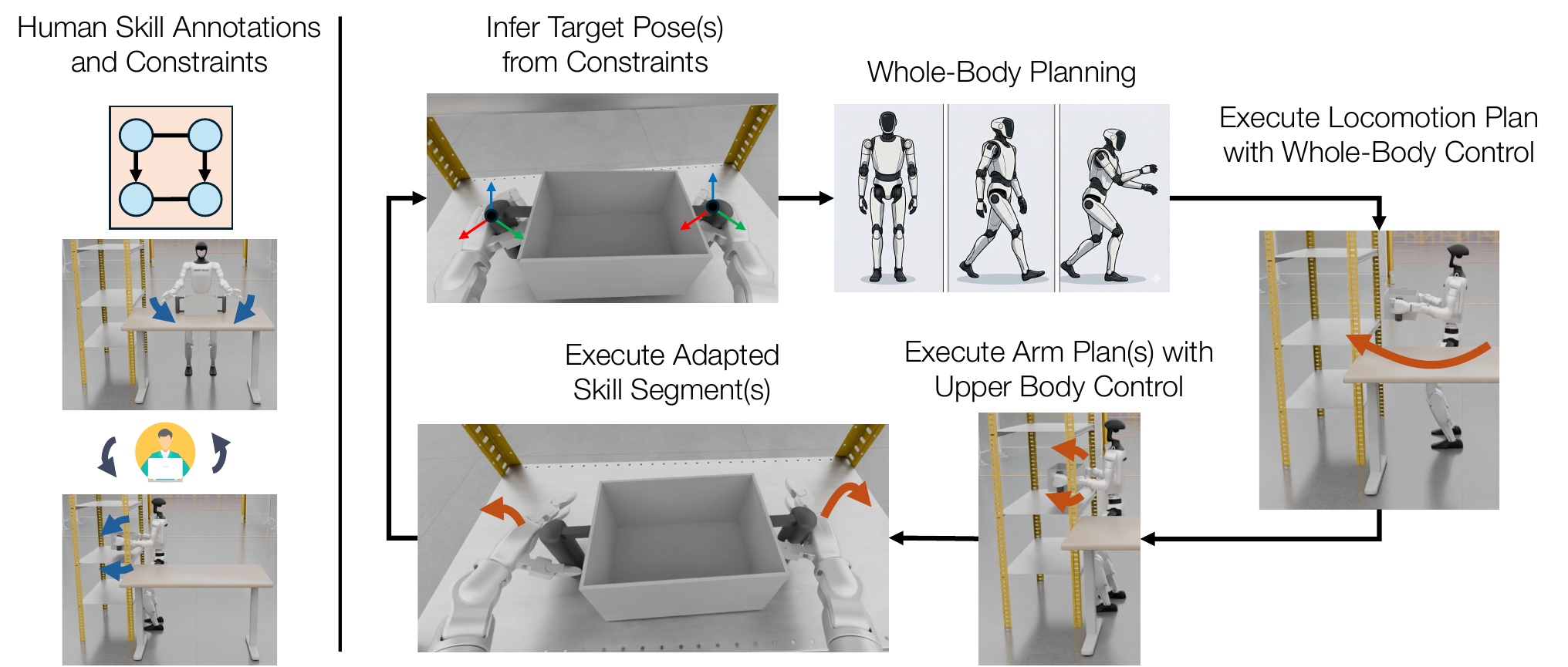}
\caption{\small{\textbf{Method Overview.} \textit{Left:} \sysName{} takes a source demonstration with per-arm skill annotations (blue arrows) and constraints (orange box). \textit{Right:} constraints determine execution order and target poses; whole-body planning produces locomotion and arm plans executed sequentially.}}
\label{fig:method}
\end{figure}

\section{Prerequisites}
\label{sec:prereq}

\textbf{Problem Statement.} We consider a bimanual legged humanoid with leg, torso, arm, and hand joints ${\cal J}=J_l\cup J_t\cup J_{a_l}\cup J_{h_l}\cup J_{a_r}\cup J_{h_r}$.
The robot state is represented by $q \in \R^{|{\cal J}|}$.
The state of each object in the world is represented by the $\SE{3}$ pose of its coordinate frame $f \in {\cal F}$.
We model each humanoid loco-manipulation task as a Partially Observable Markov Decision Process with state space ${\cal S}$, observation space ${\cal O}$, action space ${\cal A}$, and initial state distribution ${\cal S}_0$, where $\supp{\cal S}_0 \subseteq {\cal S}$.
States $s \in {\cal S}$ contain the joint positions of the robot ($s[{\cal J}] = q \in \R^{|{\cal J}|}$) as well as the world poses of the end-effector ($s[e] \in \SE{3}$) and object ($s[f] \in \SE{3}$) frames.
Observations $o \in {\cal O}$ contain the robot's proprioception as well as images from cameras on the robot.
Actions $a \in {\cal A}$ are joint-space position controller targets $a[{\cal J}] \in \R^{|{\cal J}|}$ for joints ${\cal J}$.
For notational simplicity, let $a[e] \in \SE{3}$ give the world pose of end-effector $e \in {\cal E}$ derived from its target joint positions.
For each task, we are interested in generating a dataset of $N$ demonstrations $D_N = \{d^1, ..., d^N\}$ where each demonstration $d^i$ is a sequence of state, observation, and action triplets.
Using that dataset, we train Behavior Cloning~\cite{pomerleau1989alvinn,bain1995framework} policies $\pi: {\cal O} \to {\cal A}$ that map observations to actions.
During data generation, we assume privileged state access
but we seek to train policies that only use observations.

\textbf{Data Generation via Demonstration Adaptation.} 
We assume access to a small set of annotated source demonstrations ${\cal D}_0$ and automatically amplify them into a much larger set $D_N$, where $N >> |{\cal D}_0|$, by adapting them to new initial states $s_0 \sim {\cal S}_0$~\cite{mandlekar2023mimicgen}.
We segment each demonstration $d$ into object-centric {\em skill} demonstrations $\psi = \langle e, f, d^\psi \rangle$ ~\cite{garrett2024skillgen}, where $e \in {\cal E}$ is an end-effector frame, $f \in {\cal F}$ is a reference object frame, and $d^\psi \subseteq d$ is a contiguous subsequence of $d$.
Let $\langle s^\psi_t, o_t^\psi, a^\psi_t \rangle = d^\psi[t, :]$ be the $t$th state, observation, and action on the skill demonstration.
Let $\Psi = \{\psi_1, ..., \psi_*\}$ be the set of skills segmented from the same demonstration.
We can adapt all target end-effector poses $a^\psi[e]$ on actions $a^\psi \in d^\psi[:, 2]$, to a new state $s'$ by computing:
$a'[e] = s'[f] \inv{s^\psi_0[f]} a^\psi[e].$
Intuitively, this computes the target poses of end-effector $e$ relative to the initial pose $s^\psi_0[f]$ of object frame $f$ in the skill demonstration and applies them to the current object pose $s'[f]$ in a spatially invariant manner.
We adapt and execute each skill $\psi \in \Psi$ to generate a successful demonstration.
Bimanual manipulation requires some skills to be completed {\em prior} to others or started {\em concurrently} with others~\cite{jiang2024dexmimicgen}.
We represent this with a set of precedence order pairs 
${\cal P} = \{\langle \psi_{i_1}, \psi_{i_2} \rangle, ... \}$ and concurrent order pairs ${\cal C} = \{\langle \psi_{j_1}, \psi_{j_2} \rangle, ...\}$, where 
$\langle \psi, \psi' \rangle \in {\cal P}$ indicates that skill $\psi$ must be completed before skill $\psi'$ starts and $\langle \psi, \psi' \rangle \in {\cal C}$ specifies that skills $\psi, \psi'$ must start at the same time.
For example, in Table-to-Shelf (Fig.~\ref{fig:teaser}), we define a skill per end-effector for picking the box ($f$ is the box) and placing the box ($f$ is the shelf), and impose each pick to finish before each place.

\section{HumanoidMimicGen}\label{sec:method}

Existing demonstration generation algorithms~\cite{mandlekar2023mimicgen,garrett2024skillgen,jiang2024dexmimicgen}
require that the robot's action space be in task space, where end-effector target poses are mapped to commands using, for example, Operational-Space Control (OSC)~\cite{khatib1987unified}.
However, controlling legged humanoids requires maintaining stability during both navigation and manipulation.
This requires careful whole-body coordination to balance statically and
walk dynamically. 
We thus cannot effectively apply an independent task-space control strategy to all robot limbs, notably the legs.
In \sysName{}, we address this by first adopting a {\em hybrid} action space, where the upper body is governed by a joint-space
controller, but the lower body is commanded by an RL policy trained to control the velocity of the robot's pelvis.
The RL policy can dynamically maintain stability, but not in a way that perfectly tracks velocity actions.
To compensate for this, we take a {\em decoupled} approach to data generation, where we separate planning into static manipulation and dynamic locomotion phases.
Additionally, we introduce a {\em whole-body} adaptation and planning scheme in order to replay demonstrated skills in a manner robust to changes in the environment.

Section~\ref{sec:control} describes the hybrid action space and how we use the RL lower-body controller.
Section~\ref{sec:skill} introduces a unified framework for representing skill constraints and discusses how skills are grouped and ordered.
Section~\ref{sec:pseudocode} presents the full \sysName{} data generation procedure.
Finally, Section~\ref{sec:randomization} describes perturbation strategies used to boost policy success rates.

\subsection{Humanoid Hybrid Action Space}\label{sec:control}

We use a \emph{hierarchical hybrid action space} for humanoid control \cite{ben2025homie}. Low-level joint commands use a joint-space interface, while base motion and leg coordination are handled by a learned locomotion controller.
At the lowest level, all joints ${\cal J}$
are commanded via joint position control. 
Above this interface, we expose a high-level action API with two components:
(i) joint position commands for the upper body (arms, hands, torso) and
(ii) base motion commands for locomotion.

We adopt the Homie~\cite{ben2025homie} RL locomotion controller.
Inputs include the current and target arm and torso configurations, and base command
$a[l] = [\dot{x}, \dot{y}, \dot{\theta}, z],$
where \([\dot{x}, \dot{y}]\) specify planar base velocity, \(\dot{\theta}\) specifies yaw rate, and \(z\) specifies the desired torso height.
Given this input, the RL controller produces dynamically feasible leg joint position commands, which are forwarded to the low-level joint-space API. This hierarchy enables intuitive teleoperation and data generation
while delegating balance, contact handling, and leg coordination to the learned locomotion controller.

\subsection{Skill Planning}\label{sec:skill}

From Section~\ref{sec:prereq}, we assume that a set of skills $\Psi$ for demonstration $d$ is annotated with precedence ${\cal P}$ and coordination constraints ${\cal C}$.
Because each end-effector $e$ can only perform a single skill at a time, we assume that the skills per end-effector are totally ordered in ${\cal P}$,
which means
at most $|E| = 2$ skills can be active at a time.
Because coordination constraints tie the start of two skills together, we can reduce the coordination constraint $\langle \psi, \psi'\rangle$ to additional precedence constraints by transitively applying the precedence constraints for $\psi$ to $\psi'$ and vice versa:
${\cal P} \gets {\cal P} \cup \{\langle \psi^*, \psi' \rangle \in \Psi^2 \mid \exists \psi \in \Psi. \langle \psi, \psi' \rangle \in {\cal C} \wedge \langle \psi^*, \psi \rangle \in {\cal P}\}.$
Afterward, the space of skill plans can be represented as a Directed Acyclic Graph (DAG) $\langle \Psi, {\cal P} \rangle$, where vertices are skills $\psi \in \Psi$ and directed edges $\langle \psi, \psi' \rangle \in {\cal P}$ are precedence constraints.
In Section~\ref{sec:pseudocode},
we use a greedy algorithm to produce a set of skills $\Psi_i \subseteq \Psi$ to execute next.
After planning skills $\Psi_{i-1}$, we remove them from $\Psi$ and plan skills $\Psi_i$ by selecting the remaining skills $\psi \in \Psi$ free of precedence constraints involving other skills $\psi' \in \Psi$:
$\Psi_i \gets \{\psi \in \Psi \mid \neg \exists \psi' \in \Psi. \langle \psi', \psi \rangle \in {\cal P}\}.$
This procedure resembles computing a topological sort of the DAG online and grouping incomparable skills to be executed together.
In the Table-to-Shelf task, this manifests as two iterations, where the two pick skills are executed as part of $\Psi_1$, and the two place skills are executed as part of $\Psi_2$. %
(see Fig.~\ref{fig:orders} in Appendix~\ref{app:skill}).

\subsection{Whole-Body Data Generation}\label{sec:pseudocode}

\sysName{} greedily processes applicable sets of skills $\Psi_i$ in sequence (Section~\ref{sec:skill}).
For each skill $\psi \in \Psi_i$, it computes an end-effector target pose $T[e]$ to start the skill and solves whole-body inverse kinematics to find a reachable configuration $q''$ that jointly achieves each active end-effector's target pose.
Then, it decomposes planning into a locomotion subproblem to an intermediate configuration $q'$, along with a stationary motion subproblem, ultimately to $q''$.
Finally, it performs whole-body skill adaptation to replay the skill demonstrations in the current state.
Algorithm~\ref{alg:system} in Appendix~\ref{app:pseudocode} displays the pseudocode for \proc{HumanoidMimicGen}, which executes one episode of data generation.
It takes as input an initial state $s_0$, a set of skills $\Psi$, and a set of precedence partial orders on the skills ${\cal P}$.
To generate a dataset $D_N$ of $N$ demonstrations, \proc{HumanoidMimicGen} is called repeatedly until it returns successful $N$ times, using sampled initial states $s_0 \sim {\cal S}_0$ and skills and partial order pairs $\langle \Psi, {\cal P} \rangle$ sampled from the source demonstrations.

\proc{HumanoidMimicGen} iteratively plans and executes until all skills $\psi \in \Psi$ have been completed.
On its $i$th iteration, it greedily selects skills $\Psi_i \in \Psi$ that do not have any partial order constraints involving another skill $\psi'$ that has not been completed.
For each skill $\psi \in \Psi_i$, using the current state $s$ and the first state $s^\psi_0$ on the active skill demonstrations, it adapts the demonstration pose $s_0^\psi[e]$ for end-effector $e$ using the current pose $s[f]$ of object frame $f$, resulting in target end-effector pose $T[e]$.
Next, it solves whole-body inverse kinematics to produce a batch of full configurations $Q$ that reach $T[e]$ for each active end-effector.
For each candidate target configuration $q'' \in Q$, it first computes a {\em switch} configuration $q'$, formed from the upper joint positions of $q$ and the lower joint positions of $q''$, where the robot switches from locomotion to manipulation.
Then, it plans a locomotion trajectory $\tau_l$ between the current configuration $q$ and the switch configuration $q'$. %
If planning succeeds, \proc{HumanoidMimicGen} executes $\tau_l$ using the RL controller.
If planning fails, the next configuration in the batch is attempted.
If the batch is exhausted, the episode is discarded.

Because of the imperfect execution of the RL controller, we replace $q'$ with the achieved switch configuration, plan a manipulation trajectory $\tau_m$ between the switch configuration $q'$ and target configuration $q''$, and execute $\tau_m$ using the upper-body joint-space controller.
The skill demonstration actions $a^\psi$ for each skill $\psi \in \Psi_i$ are adapted by $s[f]$ into a skill trajectory $\tau_{\Psi_i}$ using subroutine \proc{adapt-skill-demos} (Algorithm~\ref{alg:adaptation}), which solves whole-body IK to track the adapted end-effector target poses $a^\psi[e]$.
Figure~\ref{fig:adaptation} in Appendix~\ref{app:manipulation} visualizes this process applied to a single-arm pick skill on the ``Drill Lift'' task.
Hand joint positions $a^\psi[J_{h}]$ are replayed without modification.
After completing each skill in $\Psi$, the current state $s$ is checked for task success using the environment success criteria ($\proc{check-success}$).
In Appendix~\ref{app:manipulation}, we detail how we perform GPU-accelerated whole-body inverse kinematics and motion planning for humanoids in between adapted skill segments.

\subsection{Motion and initialization randomization.}
\label{sec:randomization}

During data generation, we add noise perturbations \cite{mandlekar2023mimicgen, garrett2024skillgen, jiang2024dexmimicgen, lin2025cpgen} to improve policy performance.
First, we inject motion noise during rollout: the robot executes
$a' = a + \epsilon$, where $\epsilon \sim \mathcal{N}(0, \sigma^2 I)$,
but we store the original action $a$ as the demonstration label.
This exposes the robot to off-nominal states while preserving the intended expert action.
Second, we randomly perturb the robot's initial base pose around the nominal start state
to increase initial-state diversity.

\begin{figure} %
\centering
\newcommand{\benchmarkPanel}[2]{%
\begin{minipage}[t]{0.332\linewidth}
    \centering
    \includegraphics[width=\linewidth]{figures/benchmark_tasks/#1.png}\\[-1pt]
    {\small\bfseries #2}
\end{minipage}}
\benchmarkPanel{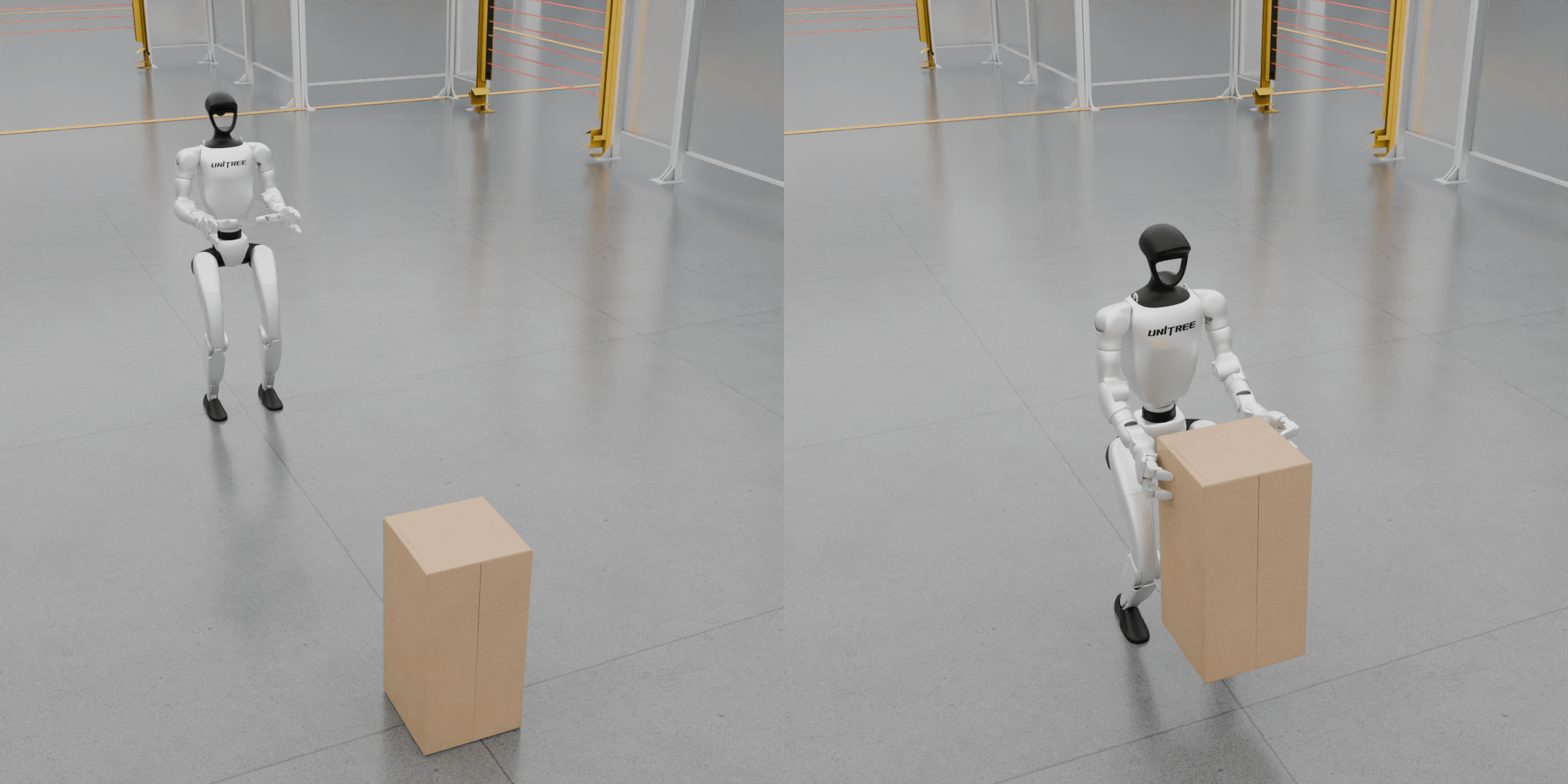}{Box Lift Floor}\hfill
\benchmarkPanel{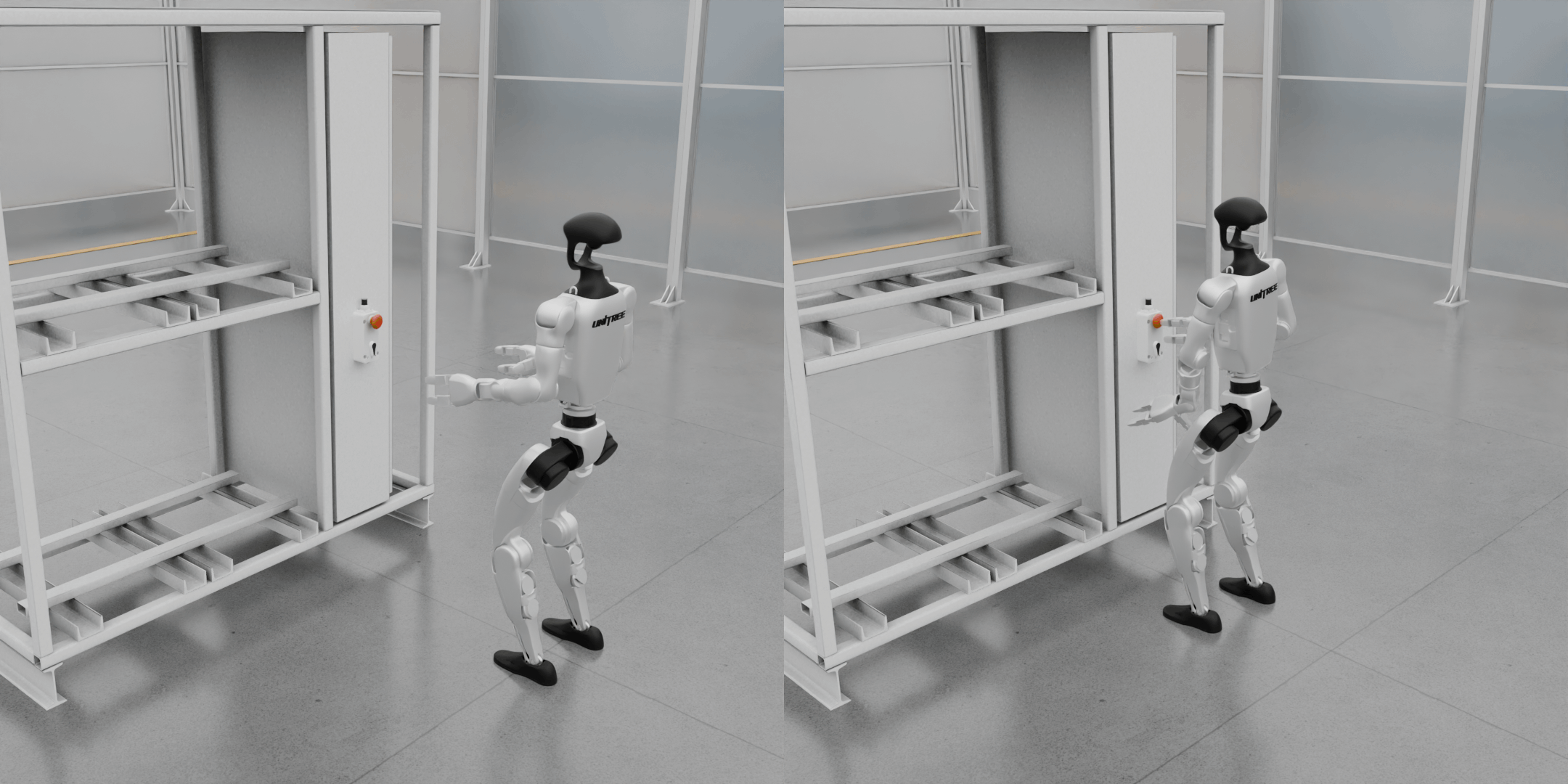}{Push Button}\hfill
\benchmarkPanel{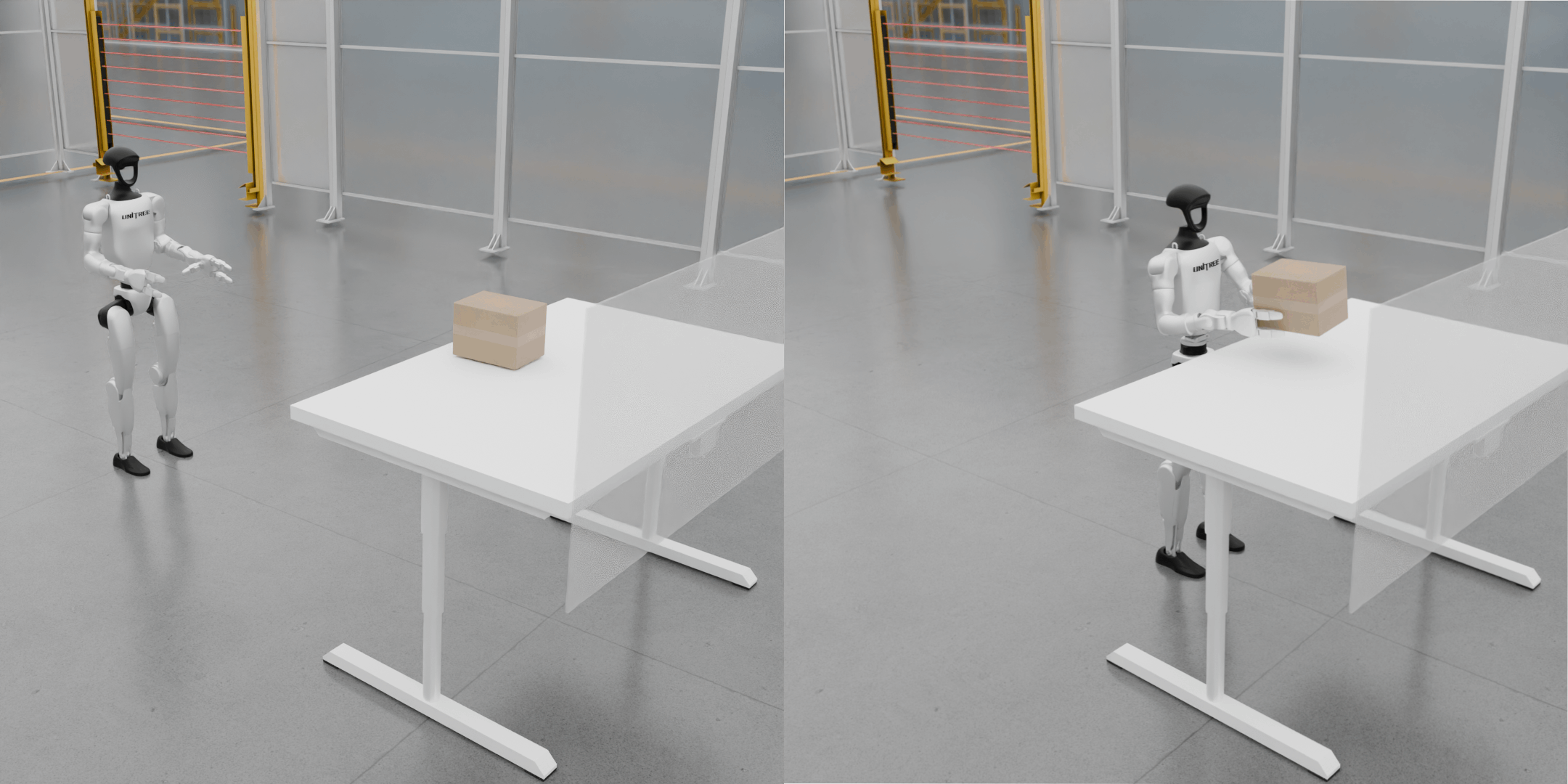}{Box Lift}
\par\vspace{5pt}
\benchmarkPanel{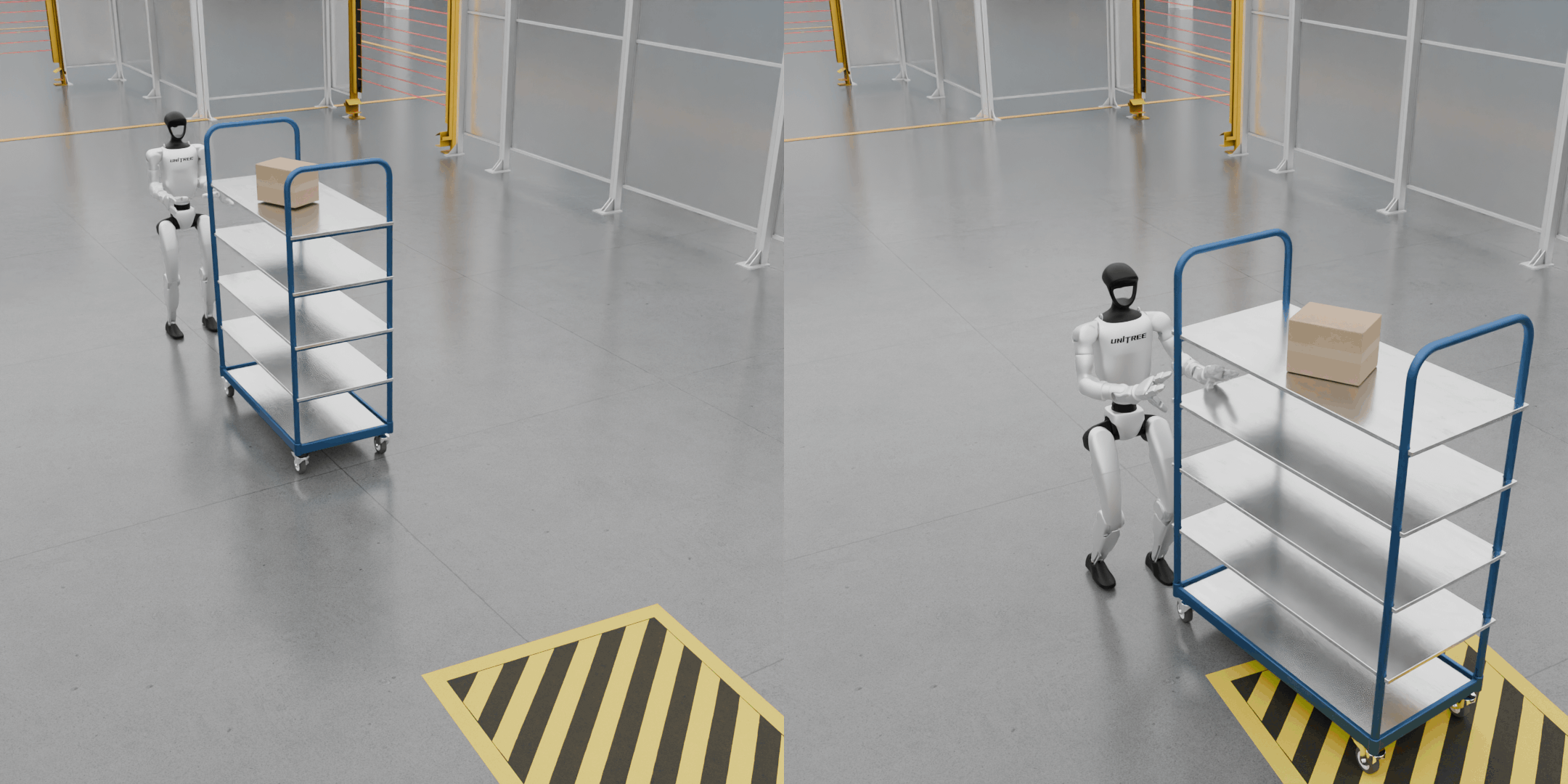}{Push Shelf Forward}\hfill
\benchmarkPanel{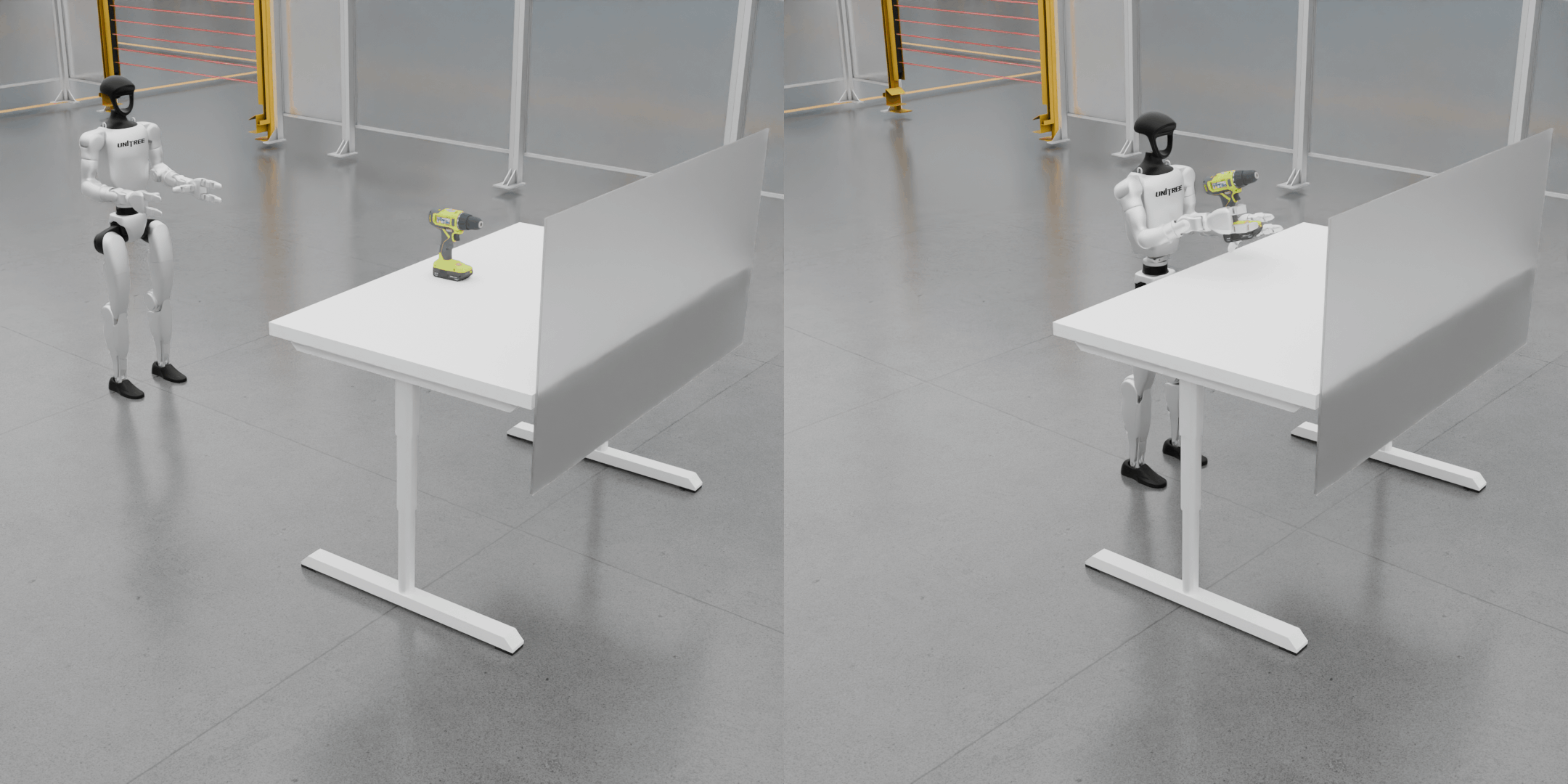}{Drill Lift}\hfill
\benchmarkPanel{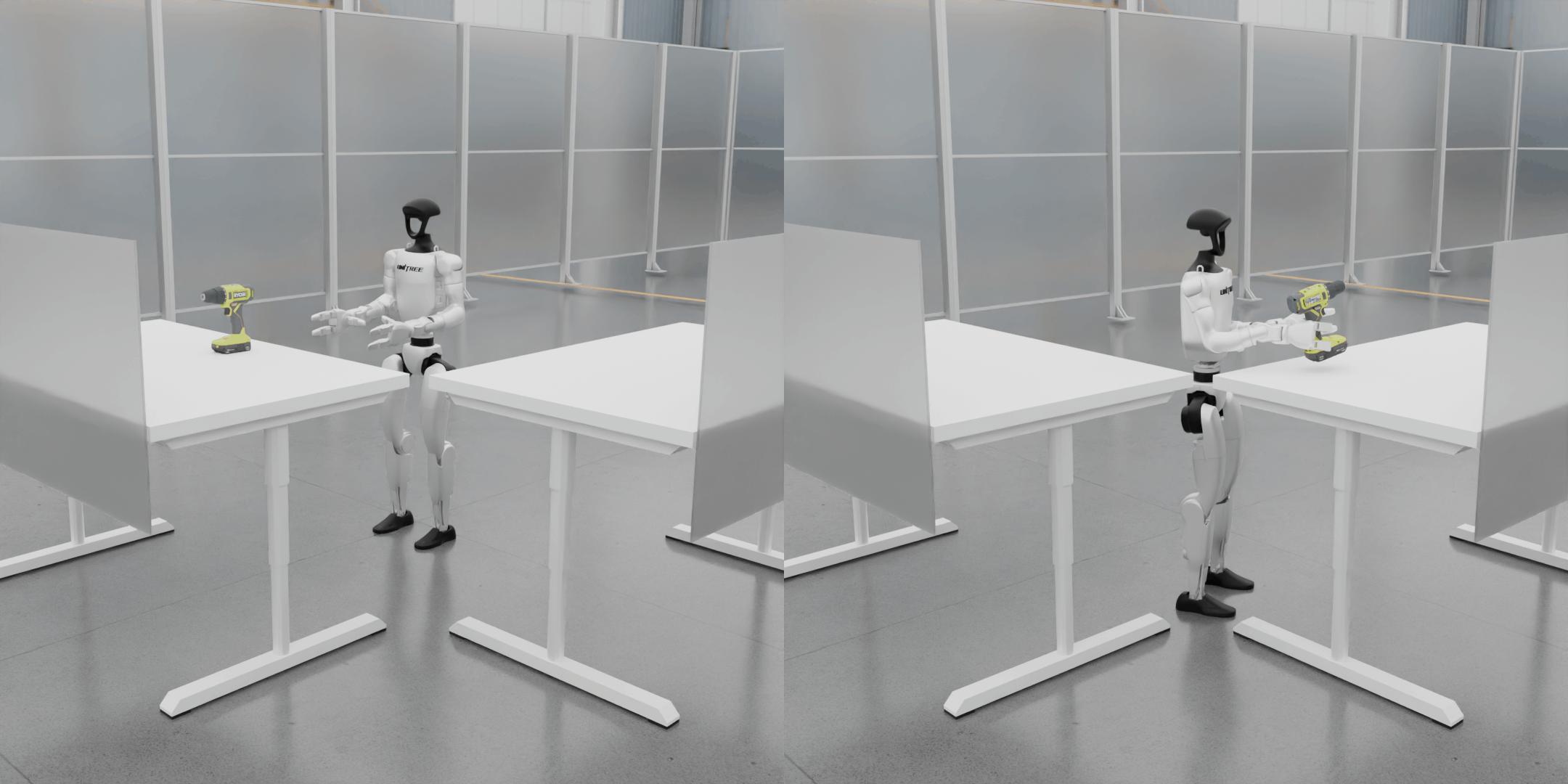}{Drill PnP}
\par\vspace{5pt}
\benchmarkPanel{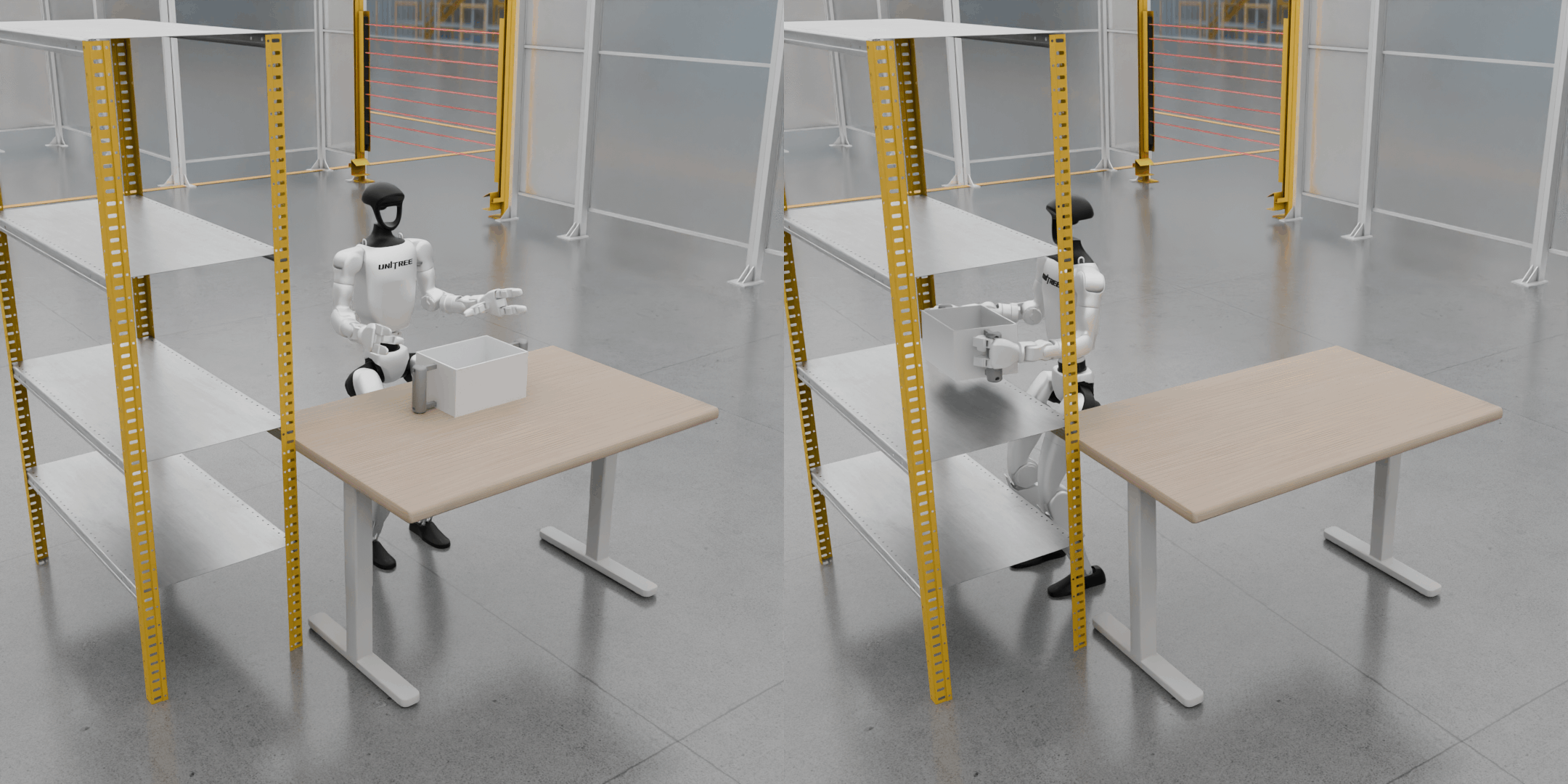}{Box Table to Shelf}\hfill
\benchmarkPanel{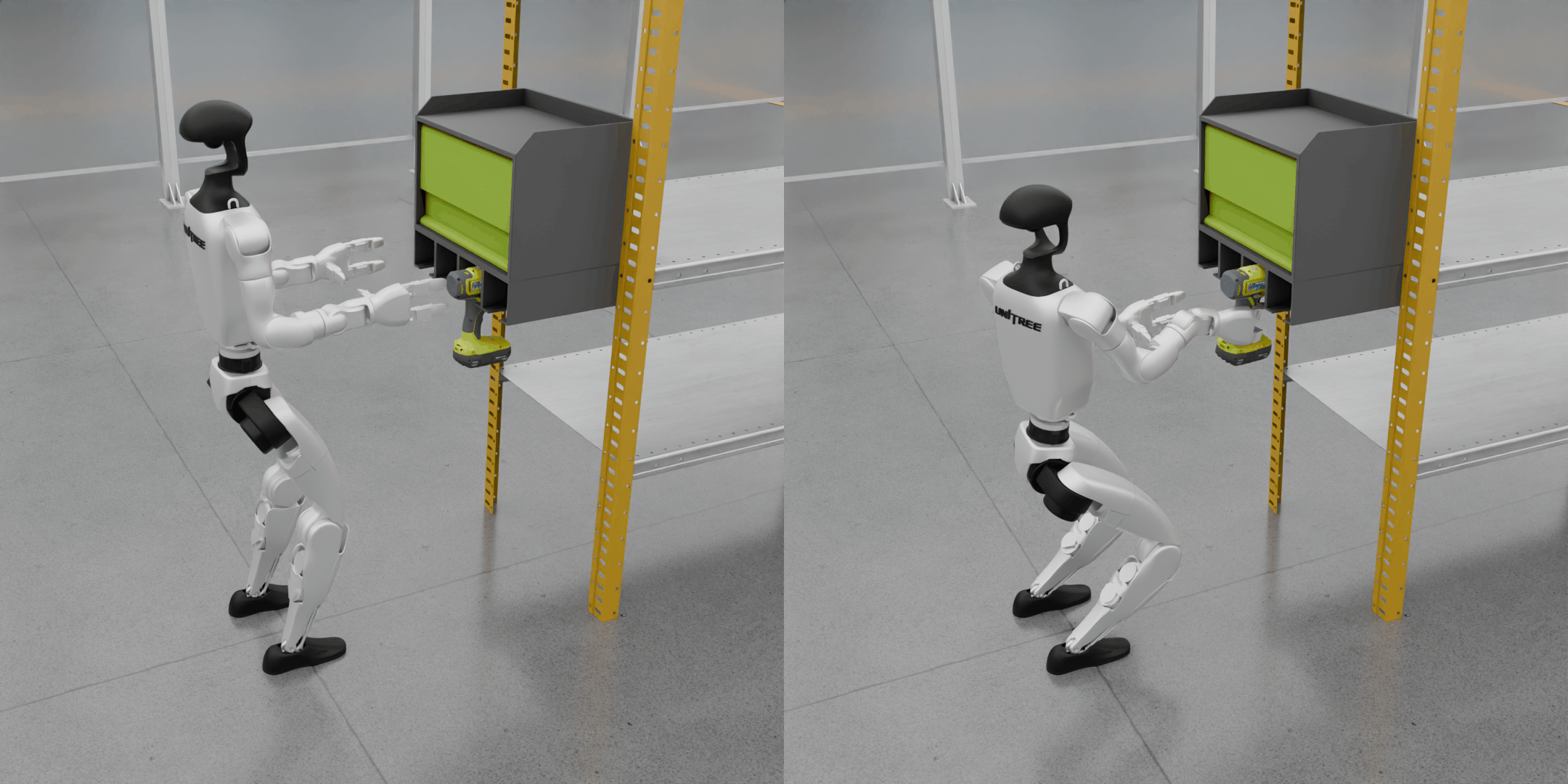}{Pick Drill From Holder}\hfill
\benchmarkPanel{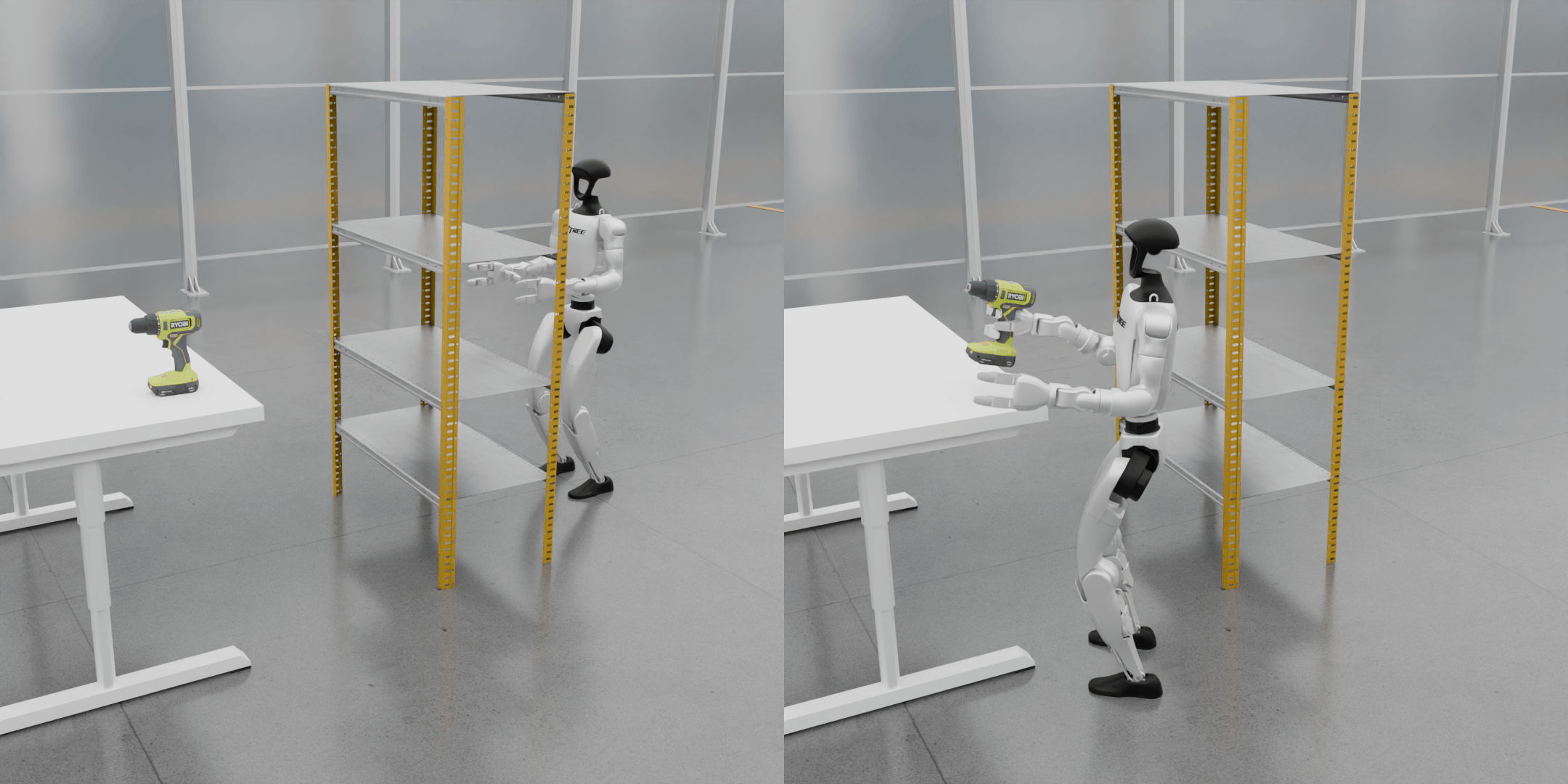}{Drill Lift Obstacle}
\caption{\small \textbf{G1 Loco-Manipulation Benchmark.} We introduce a simulation benchmark with nine loco-manipulation tasks and datasets generated by \sysName{}. Each task is shown with a sampled initial scene ({\it left}) and task-completion configuration ({\it right}). 
}
\label{fig:g1_benchmark}
\end{figure}

\section{Loco-Manipulation Simulation Benchmark}
\label{sec:benchmark}

We develop a humanoid \textbf{loco-manipulation simulation benchmark} built on robosuite~\cite{robosuite2020} and MuJoCo~\cite{todorov2012mujoco}. The benchmark tests settings where success depends on accurate base motion and whole-body coordination, enabling controlled comparisons of data generation strategies, policy architectures, and embodiment choices in loco-manipulation, as analyzed in Section~\ref{sec:experiments}.

As seen in Figure \ref{fig:g1_benchmark}, the benchmark contains \textbf{9 tasks} for a simulated G1 humanoid robot. Tasks vary along three axes: (i) required base motion, from minimal repositioning to multi-stage navigation; (ii) object interaction complexity, from single-arm to bimanual and whole-body interaction; and (iii) execution horizon. Each task uses a binary success condition. Initial states are generated by randomly sampling object poses and the robot root pose. Together, these tasks stress the locomotion--manipulation interface, where small base-placement errors can make downstream manipulation unreachable. See Appendix~\ref{app:benchmark} for more details.

\begin{table}[t]
\centering
\footnotesize
\setlength{\tabcolsep}{3.2pt}
\renewcommand{\arraystretch}{1.05}

\begin{tabular}{lcccccccccc}
\toprule
& {\makecell{Box Lift\\Floor}}
& {\makecell{Push\\Button}}
& {\makecell{Box\\Lift}}
& {\makecell{Push Shelf\\Forward}}
& {\makecell{Drill\\Lift}}
& {\makecell{Drill\\PnP}}
& {\makecell{Box Table\\to Shelf}}
& {\makecell{Pick Drill\\from Holder}}
& {\makecell{Drill Lift\\Obstacle}}
& {\makecell{Avg.}} \\
\midrule
1 Human Demo
    & 0.14 & 0.18 & 0.95 & 0.70 & 0.20 & 0.08 & 0.04 & 0.00 & 0.04 & 0.26 \\
100 Human Demos
    & 0.83 & 0.82 & 0.95 & 0.90 & 0.30 & 0.20 & 0.28 & 0.07 & 0.00 & 0.48 \\
DexMimicGen+
    & 0.87 & 0.26 & 0.68 & 0.35 & 0.13 & 0.13 & 0.17 & 0.36 & 0.00 & 0.33 \\
Ours
    & \bfseries 0.97 & \bfseries 0.92 & \bfseries 1.00 & \bfseries 1.00 & \bfseries 1.00 & \bfseries 0.70 & \bfseries 0.53 & \bfseries 1.00 & \bfseries 0.87 & \bfseries 0.89 \\
\bottomrule
\end{tabular}
\vspace{2mm}
\caption{\small \textbf{G1 Simulation Benchmark Results.} Success rates across simulated benchmark tasks for policies trained on one human source demonstration, 100 human demonstrations, DexMimicGen+ generated~\cite{jiang2024dexmimicgen} datasets, and \sysName{}-generated datasets. For DexMimicGen+ and \sysName{}, we start from the same single source demonstration for a given task and generate 1{,}000 demonstrations.}
\label{tab:main_psr_transposed}
\end{table}

\section{Experiments}
\label{sec:experiments}

We evaluate \sysName{} on the G1 loco-manipulation benchmark. We first describe the experimental setup (Sec.~\ref{subsec:implementation}) and then highlight key capabilities of \sysName{} (Sec.~\ref{subsec:features}). 
We analyze the impact of data generation and policy learning design choices (Sec.~\ref{subsec:analysis}) and finally present real-world results using sim-and-real co-training (Sec.~\ref{subsec:real_world}).

\subsection{Experimental Setup}
\label{subsec:implementation}

\textbf{Source demonstrations.}
For each task, we collect a {\em single} human demonstration using a Pico VR controller for whole-body teleoperation. 
We obtain end-effector commands from controller poses, gripper commands using controller triggers, and navigation commands with the controller thumbsticks. 
Note that the VR head-up display is not used, though the headset is used to track the controller poses.
The same teleoperation interface is used in simulation and the real world.

\textbf{Data generation.}
Starting from the single human demonstration, \sysName{} generates 1,000 successful loco-manipulation trajectories per task in simulation. 
Data generation includes randomized initial robot base poses and object poses, subject to task feasibility constraints.
For a fair comparison in Table~\ref{tab:main_psr_transposed}, DexMimicGen+ uses the same single source demonstration and the same 1{,}000-demonstration-per-task training budget.

\begin{figure*}[t]
\centering
\begin{minipage}[t]{0.49\textwidth}
    \centering
    \includegraphics[width=\linewidth]{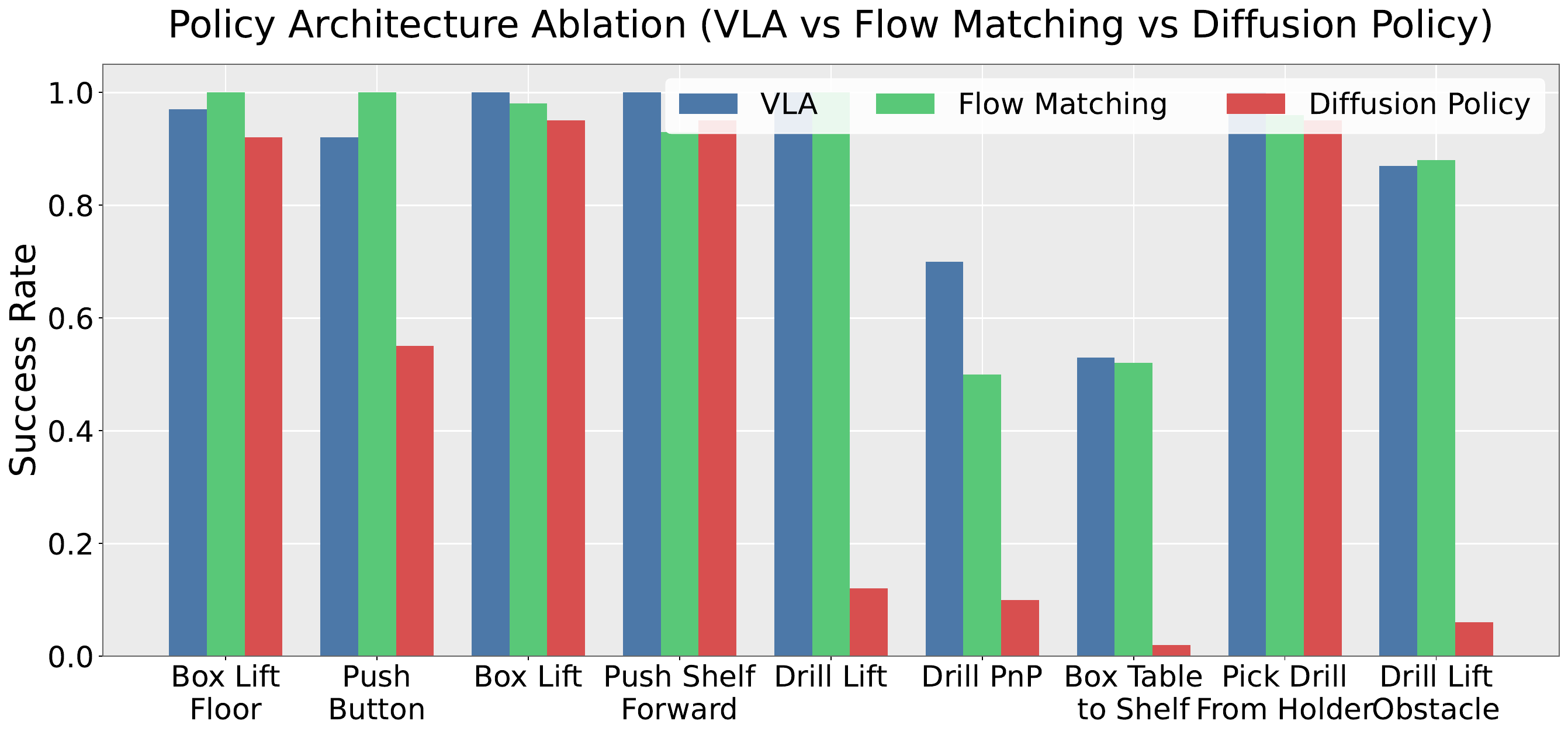}
    \centerline{\small (a) Policy architecture ablation}
\end{minipage}
\begin{minipage}[t]{0.49\textwidth}
    \centering
    \includegraphics[width=\linewidth]{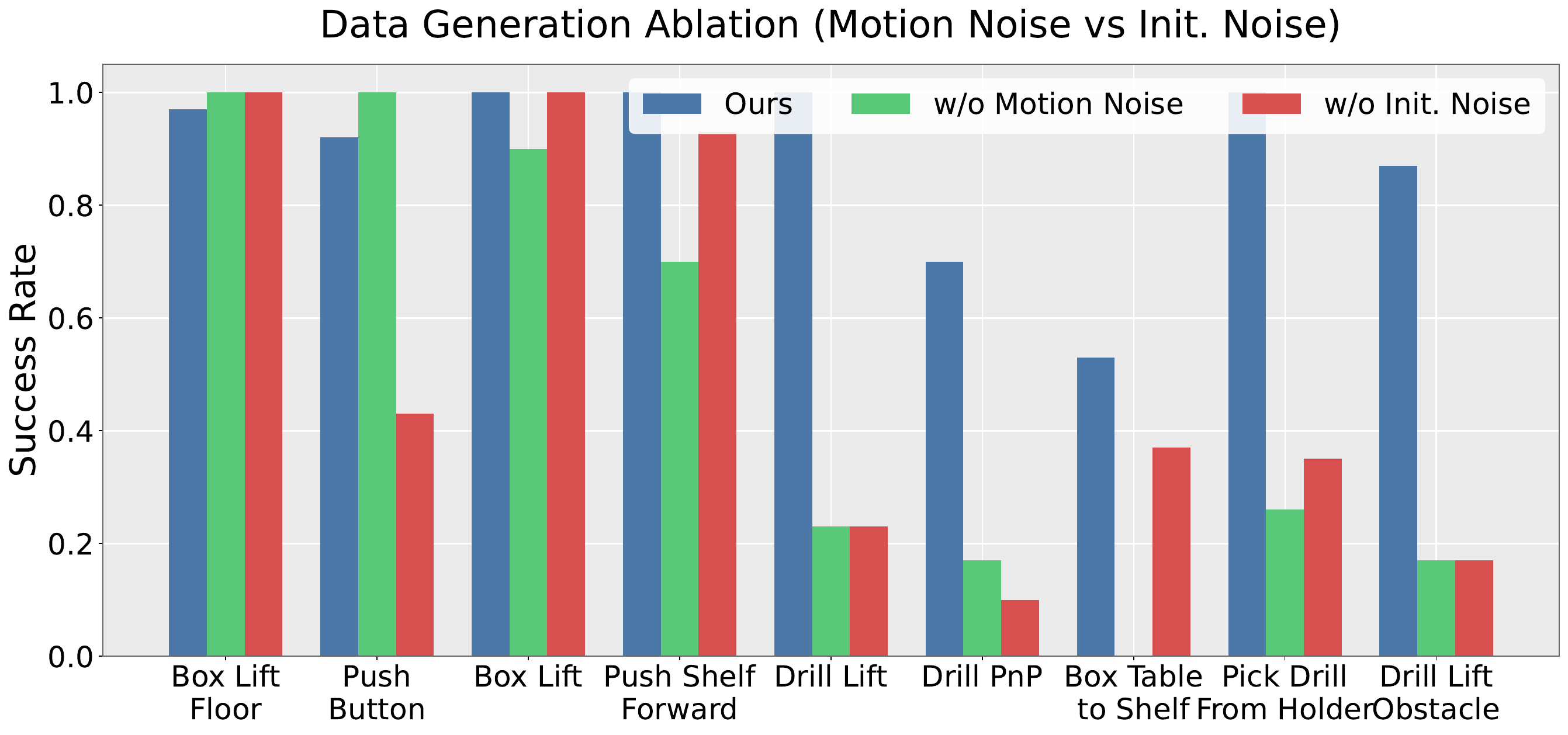}
    \centerline{\small (b) Data generation design ablation}
\end{minipage}
\caption{\small \textbf{Policy and data generation ablations.}
\textit{Left:} Policy architecture ablation. VLA outperforms Flow Matching and Diffusion Policy when all are trained on 1{,}000 \sysName{} demonstrations.
\textit{Right:} Effect of data generation design on policy success rates. Removing motion noise or initialization noise reduces policy performance, highlighting the importance of these strategies.}
\label{fig:policy_ablations_abs}
\label{fig:policy_noise_ablations}
\end{figure*}

\textbf{Policy learning.} For policy learning, we finetune a VLA model (here, the pre-trained GR00T N1.6 base model checkpoint \cite{bjorck2025gr00t}) via imitation learning. Policies take RGB observations from an ego-view camera with a resolution of (224, 224, 3) and proprioceptive state information. For each dataset, we train for 25K steps, evaluate checkpoints every 5k steps on 100 episodes, and take the best-performing checkpoint. See Appendix \ref{app:policy-training} for details.

\subsection{\sysName{} Capabilities}
\label{subsec:features}

\textbf{\sysName{} boosts policy success rates over baseline data generation methods and human demonstrations.}
We compare \sysName against policies trained purely on the single source human demonstration, 100 human demonstrations, as well as a DexMimicGen+ baseline.
The DexMimicGen+ baseline involves extending DexMimicGen~\cite{jiang2024dexmimicgen} for the loco-manipulation setting; see Appendix \ref{app:dexmimicgen-plus} for details.
From Table~\ref{tab:main_psr_transposed},  \sysName{} improves policy performance over all baselines in all tasks.
Averaged over nine tasks, data generated by \sysName{} from \textit{a single} source human demonstration increases policy performance from 0.33 (DexMimicGen+) to 0.89 and thus can reliably transform a single demonstration into diverse, high-quality training data, while baselines fail to generalize over varying states.

\textbf{\sysName{} addresses long-horizon loco-manipulation problems.}
\sysName{} generates successful demonstrations for tasks that require coordinated navigation and whole-body manipulation over longer periods of time.
Policies trained on datasets generated by \sysName{} maintain high success rates on these long-horizon tasks.
For example, \sysName{} achieves 1.00 PSR on PushShelfForward (1230 steps) and 0.97 on BoxLiftFloor (900 steps), compared to 0.35 and 0.87 with DexMimicGen+, and 0.70 and 0.14 with source demonstrations alone.

\subsection{\sysName{} Analysis}
\label{subsec:analysis}

\textbf{Effect of motion noise and robot pose initialization randomization.}
Figure~\ref{fig:policy_noise_ablations} compares datasets generated with synthetic motion randomization and randomized robot base initialization (described in Section \ref{sec:randomization}) against counterparts without these perturbations.
Removing motion noise reduces mean success from 0.89 to 0.49, while fixing initial robot pose reduces success from 0.89 to 0.51.
These degradations are consistent across most tasks, showing that both trajectory and state-level randomization improve policy success rates.

\textbf{Effect of different policy architectures trained on datasets generated by \sysName{}.}
We compare our VLA model finetuned on data generated by \sysName{} with a flow-matching transformer policy following AdaFlow~\cite{hu2024adaflow} and a Diffusion Policy baseline~\cite{chi2023diffusion} on the G1 loco-manipulation benchmark. 
In Figure~\ref{fig:policy_ablations_abs}, we see the finetuned VLA achieves the highest average PSR at 0.89, slightly outperforming flow matching at 0.86, while both outperform diffusion policy at 0.51.

\subsection{Real-World Evaluation}
\label{subsec:real_world}

\begin{figure}[t]
\centering

\begin{minipage}[t]{0.49\columnwidth}
\centering
\includegraphics[width=0.5\linewidth]{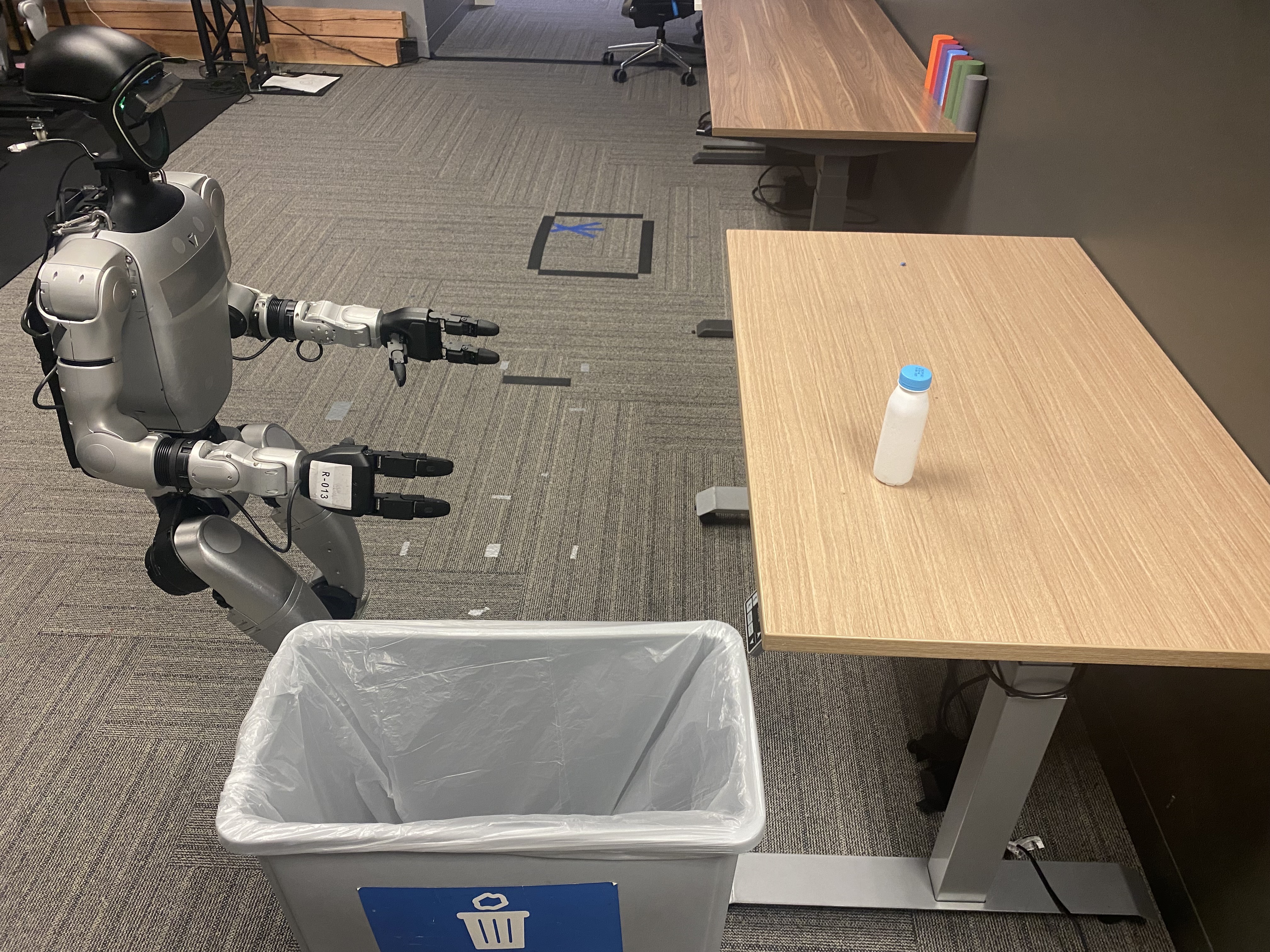}%
\includegraphics[width=0.5\linewidth]{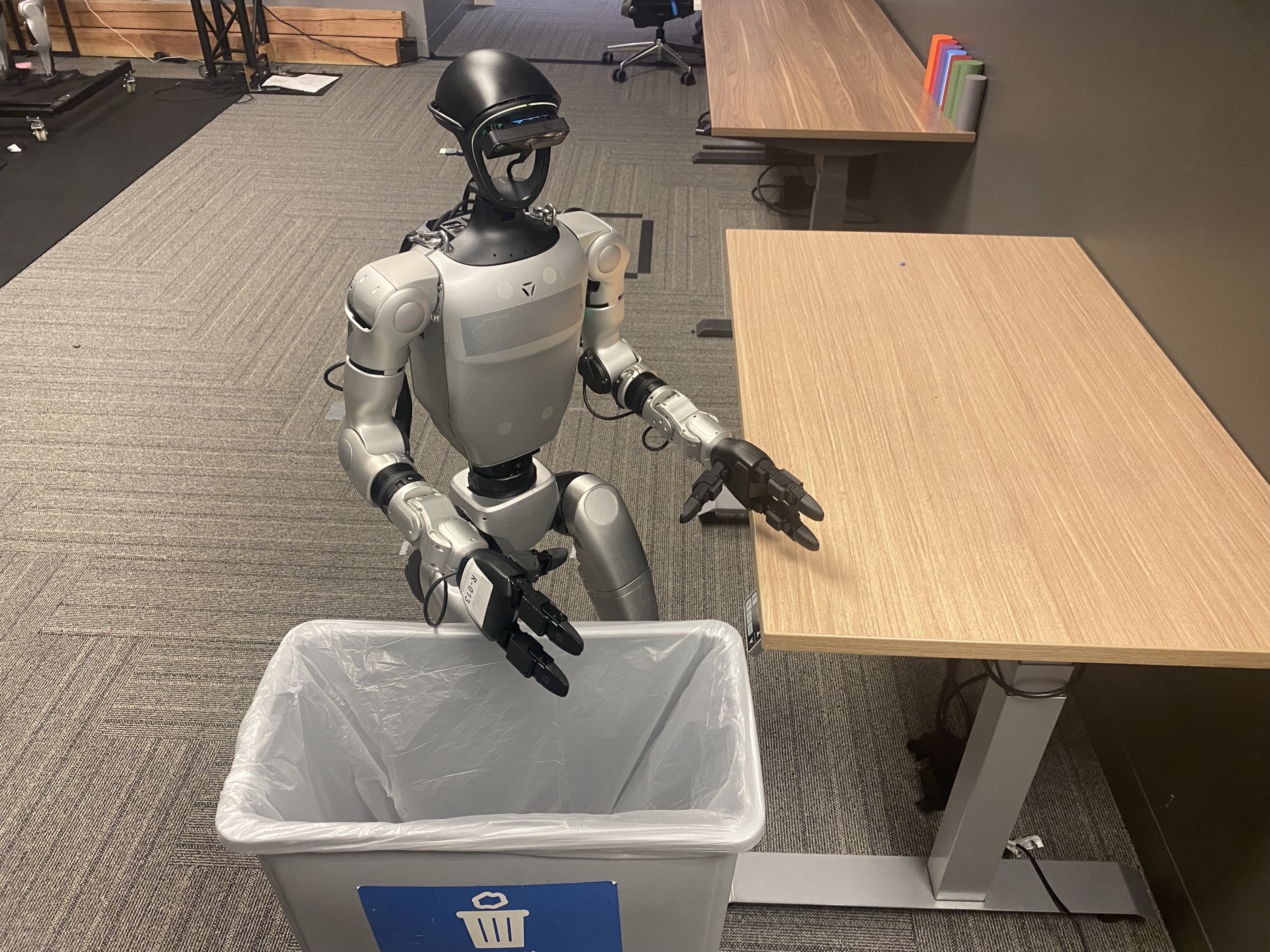}
\end{minipage}
\hfill
\begin{minipage}[t]{0.49\columnwidth}
\centering
\includegraphics[width=0.5\linewidth]{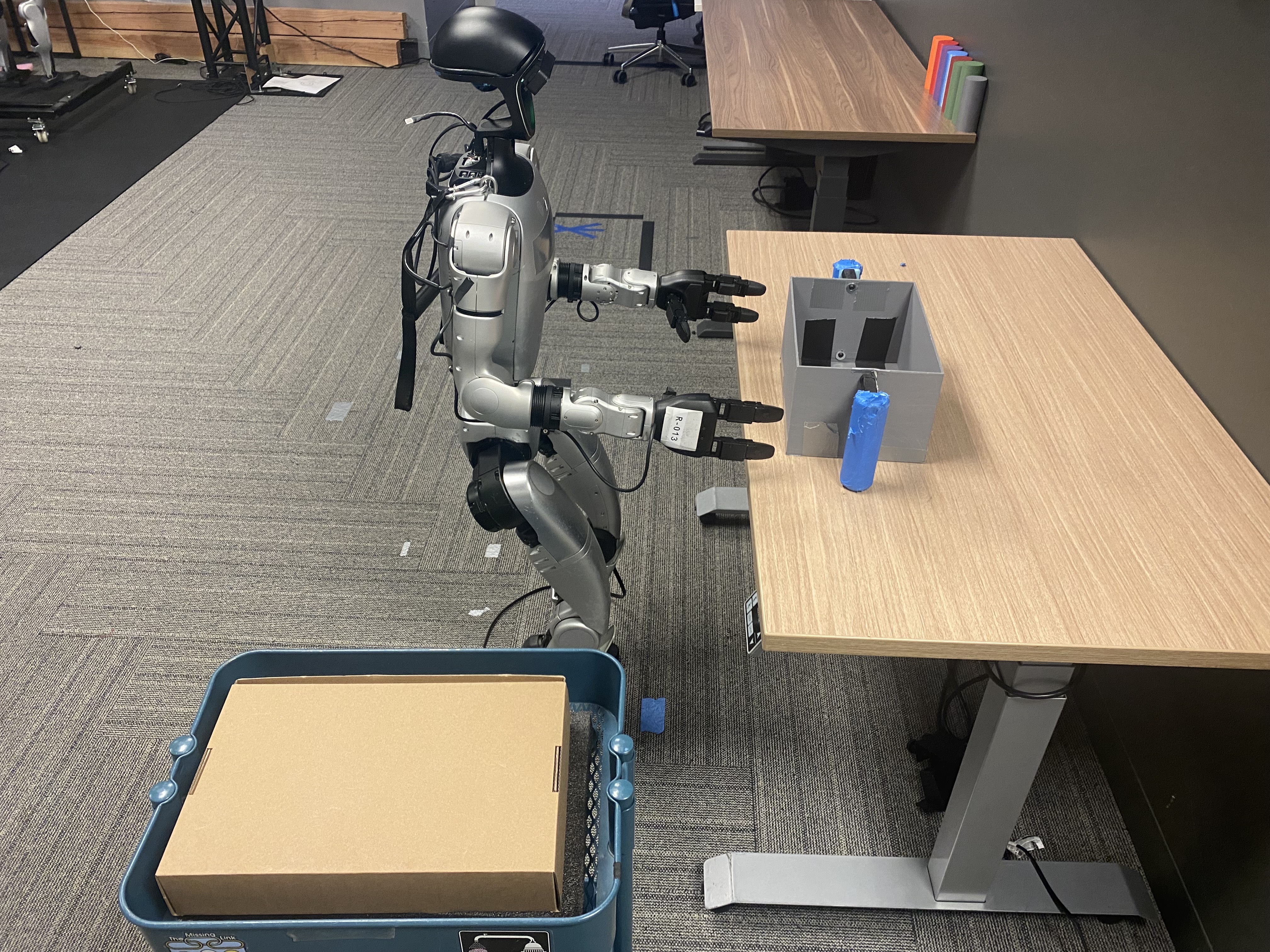}%
\includegraphics[width=0.5\linewidth]{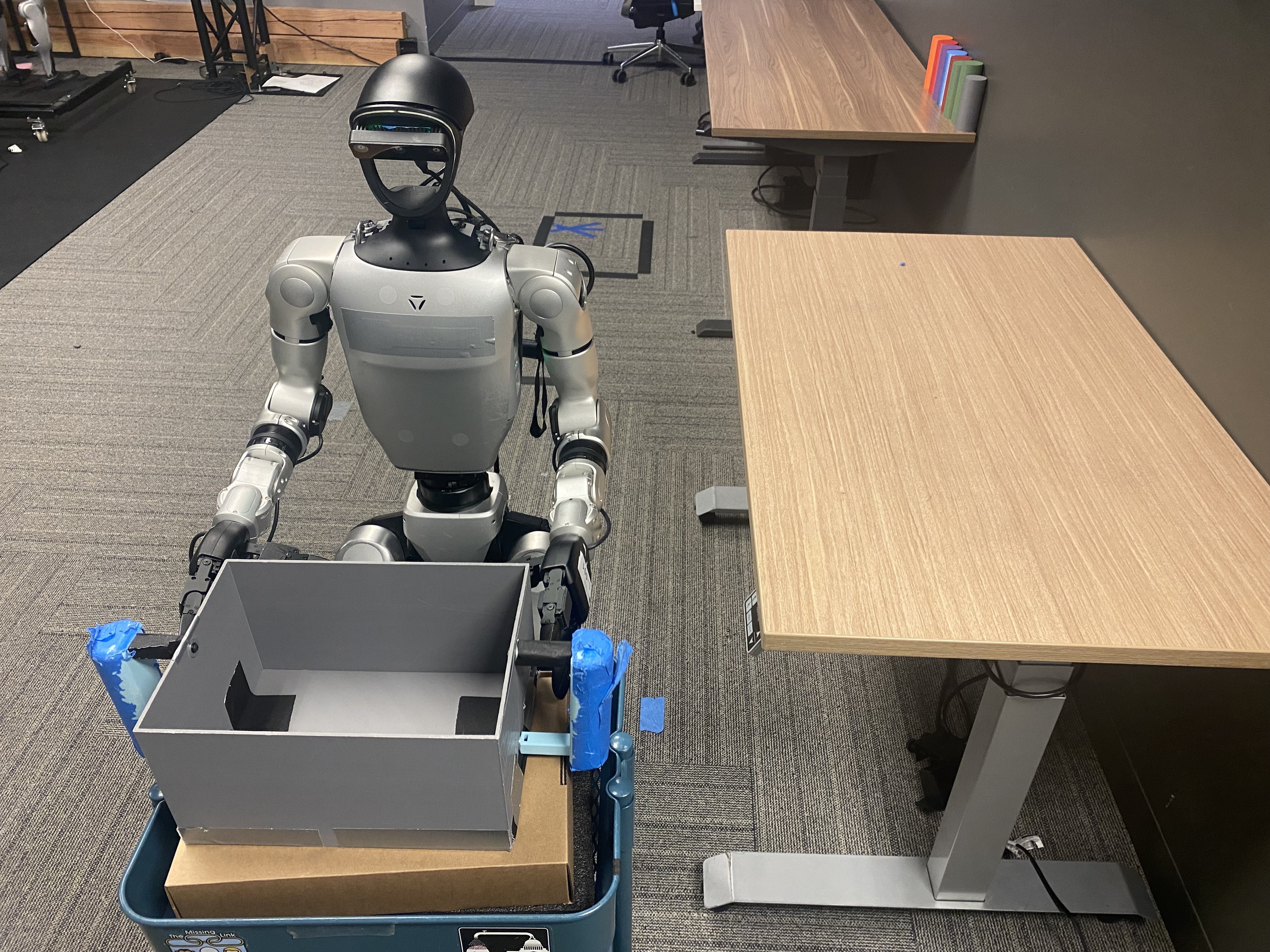}
\end{minipage}

\vspace{1pt}

\begin{minipage}[t]{0.49\columnwidth}
\centering
\includegraphics[width=0.5\linewidth]{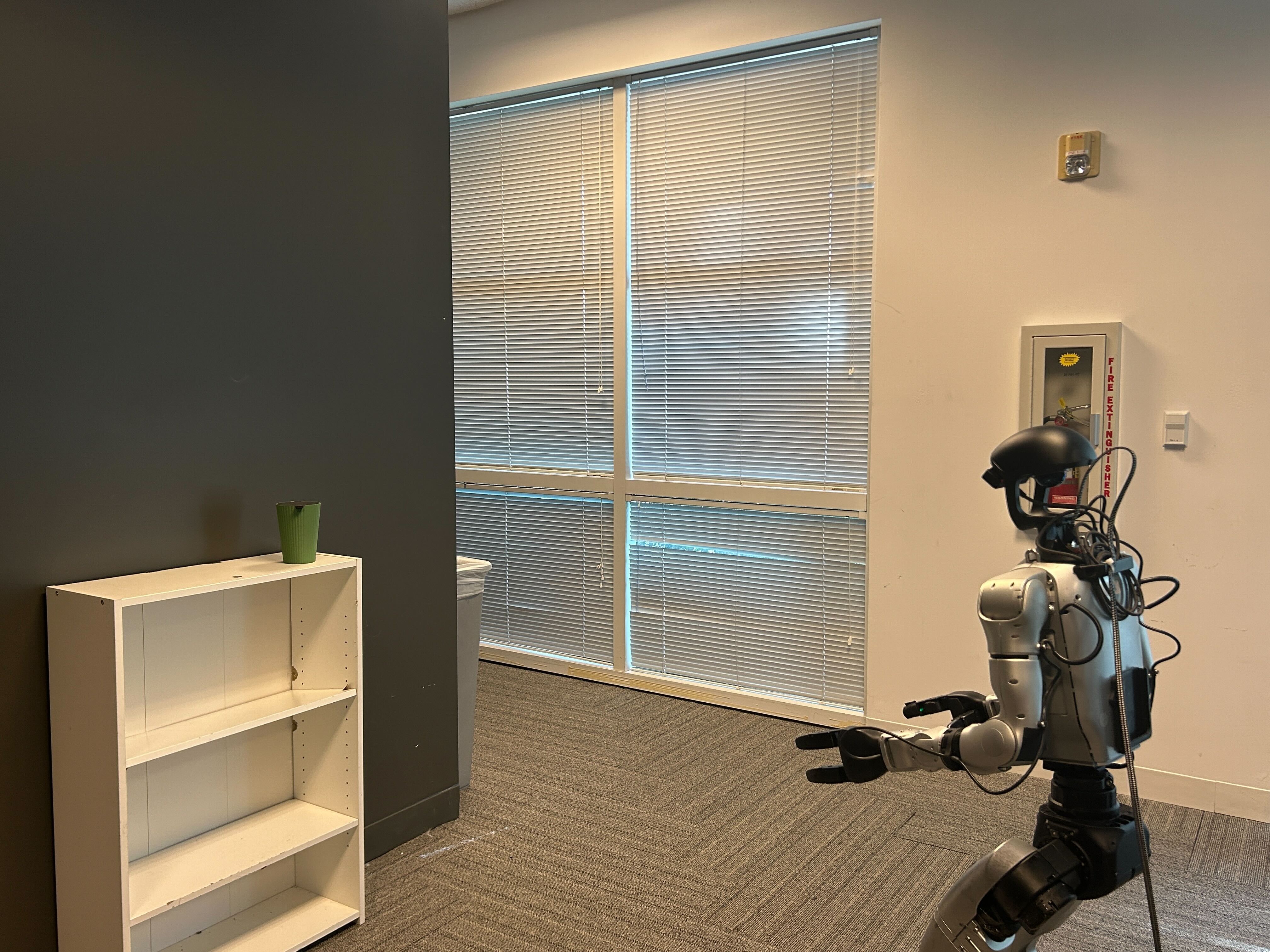}%
\includegraphics[width=0.5\linewidth]{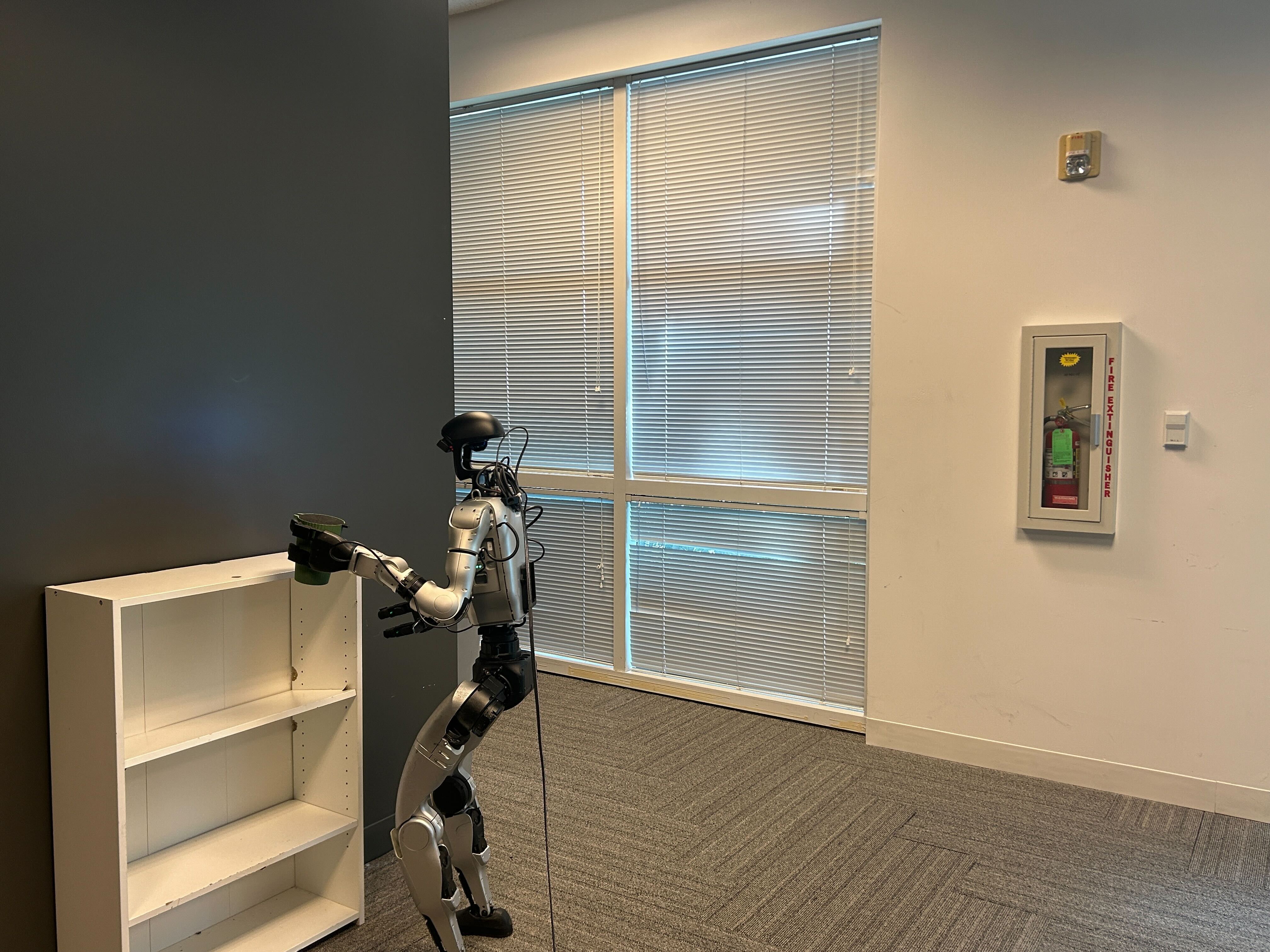}
\end{minipage}
\hfill
\begin{minipage}[t]{0.49\columnwidth}
\centering
\includegraphics[width=0.5\linewidth]{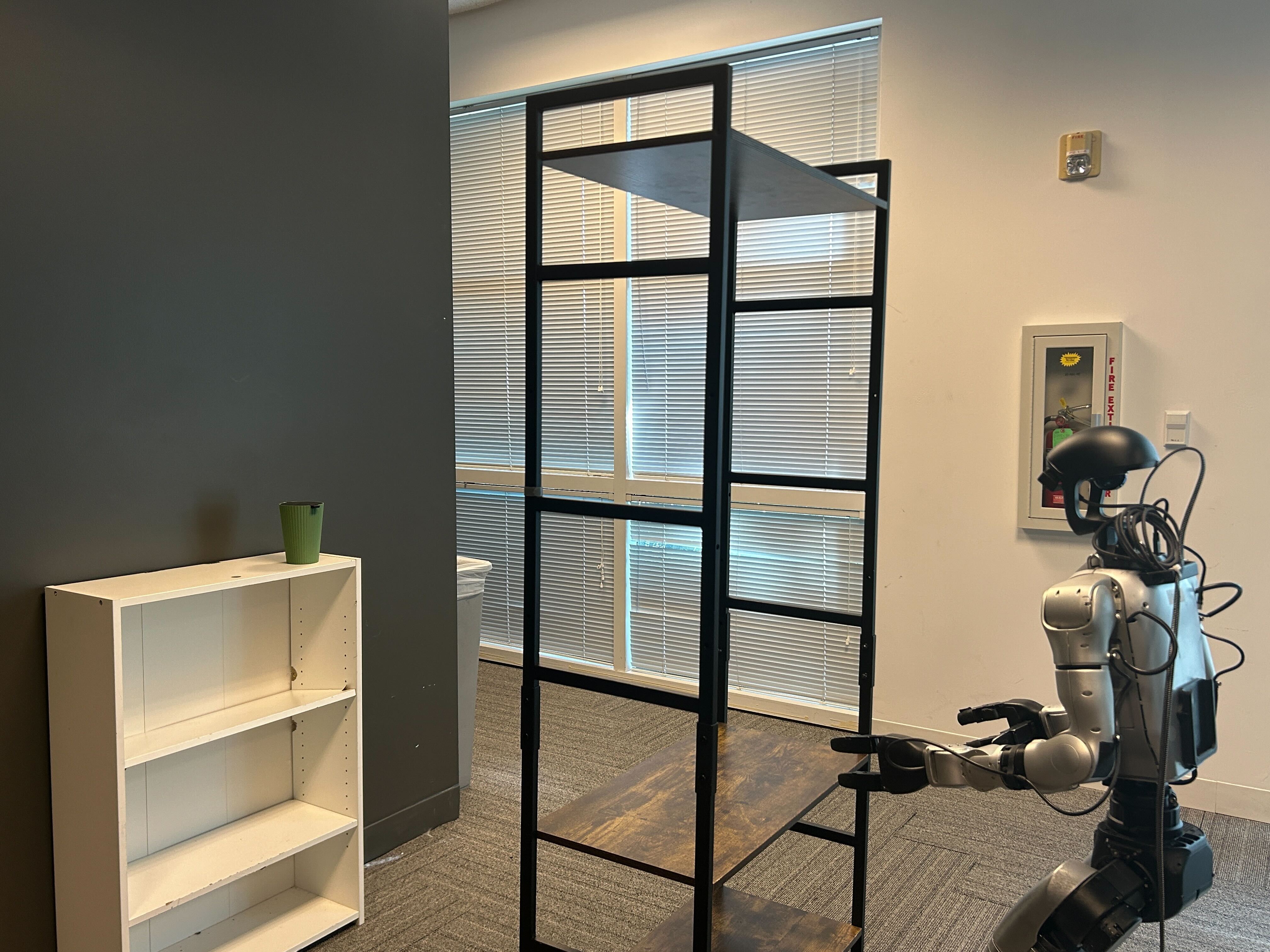}%
\includegraphics[width=0.5\linewidth]{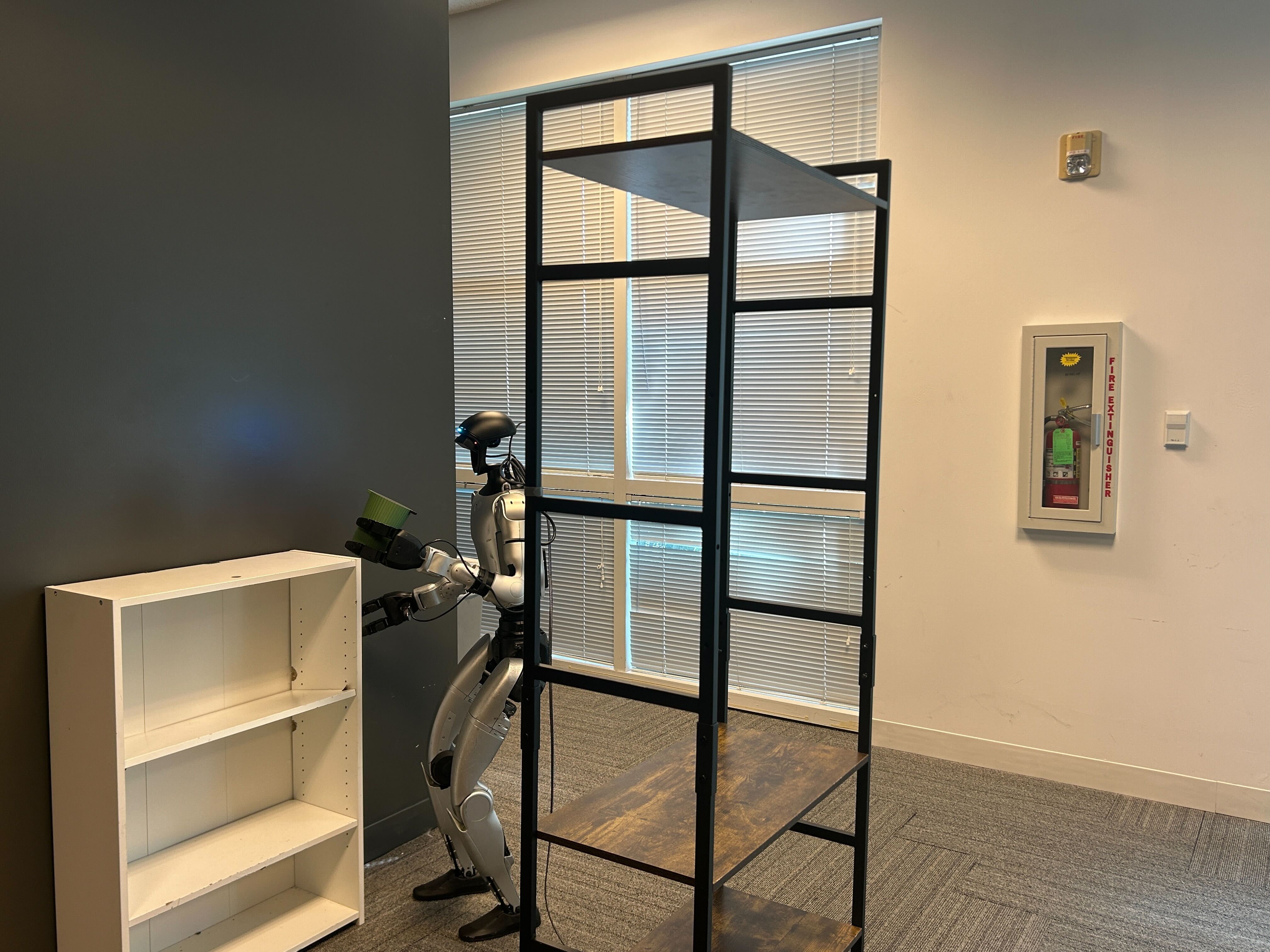}
\end{minipage}

\vspace{4pt}

{\small
\renewcommand{\arraystretch}{0.95}
\begin{tabular*}{\columnwidth}{@{\extracolsep{\fill}}lccccc@{}}
\toprule
& ThrowBottle
& BoxToCart
& PickCanister
& PickCanisterWithObstruction
& Avg. \\
\midrule
Real Only   & 0.60 & 0.35 & 0.50 & 0.60 & 0.51 \\
Co-training & \textbf{0.75} & \textbf{0.60} & \textbf{0.75} & \textbf{0.75} & \textbf{0.71} \\
\bottomrule
\end{tabular*}
}

\caption{\small \textbf{Real-World Deployment.}
We evaluate policies co-trained on \sysName simulation data and real-world demonstrations across four real-world manipulation tasks: ThrowBottle ({\it top left}), BoxToCart ({\it top right}), PickCanister ({\it bottom left}), and PickCanisterWithObstruction ({\it bottom right}).
Each left image is the initial state and right image is a goal state.
Co-training improves average policy score from 0.51 to 0.71. 
}
\label{fig:real_world_tasks}
\end{figure}

We evaluate \sysName{} on a real-world G1 humanoid robot using sim-and-real co-trained policies and compare them with policies trained only on teleoperated real robot data, following standard sim-and-real co-training practice~\cite{maddukuri2025sim, wei2025empirical}. See Appendix \ref{app:real-world} for details.
For each task, we collect a single simulation demo and automatically generate 500 more using \sysName{}.

We evaluate flow-matching policies trained from scratch on \textbf{BoxToCart}, \textbf{ThrowBottle}, \textbf{PickCanister} and \textbf{PickCanisterWithObstruction} tasks as seen in Figure~\ref{fig:real_world_tasks}. \textbf{BoxToCart} involves picking up a box with two hands, walking to a cart, then placing the box on the cart. %
\textbf{ThrowBottle} involves walking to a table, picking up a bottle, turning the waist, and throwing it into the bin.
For these two tasks, we use 30 real world and 500 sim-generated demonstrations. For \textbf{PickCanister} and \textbf{PickCanisterWithObstruction}, we train a single flow matching policy on 50 real-world demos of each variant and 1000 sim-generated demonstrations. See Appendix \ref{app:real-world-scoring} for the partial scoring criteria. 

As shown in Figure~\ref{fig:real_world_tasks}, policies co-trained on simulation and real-world data consistently outperform real-only policies, with gains of $0.15$ on \textbf{ThrowBottle}, $0.25$ on \textbf{BoxToCart}, $0.25$ on \textbf{PickCanister}, and $0.15$ on \textbf{PickCanisterWithObstruction}. These results demonstrate that data generated by \sysName{} improves real-world loco-manipulation performance.

\section{Limitations}
\sysName{} requires manual annotation of skill segments and precedence or coordination constraints, similar to DexMimicGen~\cite{jiang2024dexmimicgen} and SkillGen~\cite{garrett2024skillgen}. 
The co-training simulation environments, initial-state distributions, and success conditions are designed manually. Future work could automate environment construction and skill and constraint annotation using foundation models. 
Our method also assumes a fixed set of object-centric skills and a fixed skill sequence structure; this limits generalization to tasks requiring new skill sequences or high-level plans. 
Because \sysName{} also relies on rigid object-frame transformations, it does not currently handle large intra-category geometric variation or ambiguous contact affordances, which could be addressed by methods such as CP-Gen~\cite{lin2025cpgen}.

\section{Conclusion}

We introduce \sysName{}, a method for generating demonstrations for humanoid robot loco-manipulation, where we use a small number of expert demonstrations and compose navigation, whole-body motion, and manipulation segments from the original demonstrations. 
We also introduce the G1 Loco-Manipulation benchmark, a nine-task simulation benchmark suite designed for whole-body loco-manipulation. 
We validate the effectiveness of data generated by \sysName{} for real-world policy performance via sim-and-real co-training, where generated simulation data improves real humanoid performance when combined with limited real demonstrations.

\section*{Acknowledgments}

We thank Dennis Da, Fernando Castaneda Garcia-Rozas, Scott Reed, You Liang Tan, Fengyuan Hu, Jimmy Wu, Mengda Xu, Kaushil Prakashbhai Kundalia, Avnish Narayan, Letian (Max) Fu and Qi Wang for helpful discussions and feedback.

We thank Peter Pham, Smit Patel, Amitoj Sandhu, Amanpreet Singh, Leilee Naderi, Ethan Sick, Khushboo Gupta, Krishna Nathani, Jazmin Sanchez, Demetria Quijada, and Jesse Yang for their support with data collection and feedback on the teleoperation user experience. We also thank Jeremy Chimienti and Tri Cao for ensuring robot readiness and performing repairs.

\bibliographystyle{plainnat}
\bibliography{references}

@article{garrett2021integrated,
    title={Integrated task and motion planning},
    author={Garrett, Caelan Reed and Chitnis, Rohan and Holladay, Rachel and Kim, Beomjoon and Silver, Tom and Kaelbling, Leslie Pack and Lozano-P{\'e}rez, Tom{\'a}s},
    journal={Annual review of control, robotics, and autonomous systems},
    volume={4},
    pages={265--293},
    year={2021},
    publisher={Annual Reviews}
}

@incollection{Hauser2011Multi-modalTask,
    title = {{Multi-modal motion planning for a humanoid robot manipulation task}},
    year = {2011},
    booktitle = {Robotics Research},
    author = {Hauser, Kris and Ng-Thow-Hing, Victor and Gonzalez-Ba{\~{n}}os, Hector},
    pages = {307--317},
    publisher = {Springer}
}

@article{HauserIJRR11,
    title = {{Randomized multi-modal motion planning for a humanoid robot manipulation task}},
    year = {2011},
    journal = {International Journal of Robotics Research (IJRR)},
    author = {Hauser, Kris and Ng-Thow-Hing, Victor},
    number = {6},
    pages = {676--698},
    volume = {30},
    publisher = {Springer},
}

@inproceedings{dalibard2010manipulation,
    title = {Manipulation of documented objects by a walking humanoid robot},
    author = {Dalibard, S{\'e}bastien and Nakhaei, Alireza and Lamiraux, Florent and Laumond, Jean-Paul},
    booktitle = {2010 10th IEEE-RAS International Conference on Humanoid Robots},
    pages = {518--523},
    year = {2010},
    organization = {IEEE},
    doi = {10.1109/ICHR.2010.5686827}
}

@article{dalibard2013dynamic,
    title = {Dynamic walking and whole-body motion planning for humanoid robots: an integrated approach},
    author = {Dalibard, S{\'e}bastien and El Khoury, Antonio and Lamiraux, Florent and Nakhaei, Alireza and Ta{\"i}x, Michel and Laumond, Jean-Paul},
    journal = {The International Journal of Robotics Research},
    volume = {32},
    number = {9-10},
    pages = {1089--1103},
    year = {2013},
    doi = {10.1177/0278364913481250}
}

@inproceedings{burget2013wholebody,
    title = {Whole-body motion planning for manipulation of articulated objects},
    author = {Burget, Felix and Hornung, Armin and Bennewitz, Maren},
    booktitle = {2013 IEEE International Conference on Robotics and Automation},
    pages = {1656--1662},
    year = {2013},
    organization = {IEEE},
    doi = {10.1109/ICRA.2013.6630792}
}

@inproceedings{grey2016humanoid,
  title={Humanoid manipulation planning using backward-forward search},
  author={Grey, Michael X and Garrett, Caelan R and Liu, C Karen and Ames, Aaron D and Thomaz, Andrea L},
  booktitle={2016 IEEE/RSJ International Conference on Intelligent Robots and Systems (IROS)},
  pages={5467--5473},
  year={2016},
  organization={IEEE}
}

@article{asif2019wholebody,
    title = {Whole-body motion and footstep planning for humanoid robots with multi-heuristic search},
    author = {Asif, Rizwan and Athar, Ali and Mehmood, Faisal and Islam, Fahad and Ayaz, Yasar},
    journal = {Robotics and Autonomous Systems},
    volume = {116},
    pages = {51--63},
    year = {2019},
    doi = {10.1016/j.robot.2019.03.007}
}

@article{murooka2021humanoid,
    title = {Humanoid loco-manipulation planning based on graph search and reachability maps},
    author = {Murooka, Masaki and Kumagai, Iori and Morisawa, Mitsuharu and Kanehiro, Fumio and Kheddar, Abderrahmane},
    journal = {IEEE Robotics and Automation Letters},
    volume = {6},
    number = {2},
    pages = {1840--1847},
    year = {2021},
    doi = {10.1109/LRA.2021.3060728}
}

@article{ferrari2023multicontact,
    title = {Multi-contact planning and control for humanoid robots: Design and validation of a complete framework},
    author = {Ferrari, Paolo and Rossini, Luca and Ruscelli, Francesco and Laurenzi, Arturo and Oriolo, Giuseppe and Tsagarakis, Nikos G. and Mingo Hoffman, Enrico},
    journal = {Robotics and Autonomous Systems},
    volume = {166},
    pages = {104448},
    year = {2023},
    doi = {10.1016/j.robot.2023.104448}
}

@inproceedings{mcdonald2022guided,
    title={Guided imitation of task and motion planning},
    author={McDonald, Michael James and Hadfield-Menell, Dylan},
    booktitle={Conference on Robot Learning},
    pages={630--640},
    year={2022},
    organization={PMLR}
}

@article{dalal2023imitating,
    title={Imitating task and motion planning with visuomotor transformers},
    author={Dalal, Murtaza and Mandlekar, Ajay and Garrett, Caelan and Handa, Ankur and Salakhutdinov, Ruslan and Fox, Dieter},
    journal={arXiv preprint arXiv:2305.16309},
    year={2023}
}

@inproceedings{mandlekar2023hitltamp,
    title={Human-In-The-Loop Task and Motion Planning for Imitation Learning},
    author={Mandlekar, Ajay and Garrett, Caelan and Xu, Danfei and Fox, Dieter},
    booktitle={7th Annual Conference on Robot Learning},
    year={2023}
}

@inproceedings{
    mandlekar2023mimicgen,
    title={MimicGen: A Data Generation System for Scalable Robot Learning using Human Demonstrations},
    author={Ajay Mandlekar and Soroush Nasiriany and Bowen Wen and Iretiayo Akinola and Yashraj Narang and Linxi Fan and Yuke Zhu and Dieter Fox},
    booktitle={7th Annual Conference on Robot Learning},
    year={2023},
    url={https://openreview.net/forum?id=dk-2R1f_LR}
}

@article{jiang2024dexmimicgen,
    title={Dexmimicgen: Automated data generation for bimanual dexterous manipulation via imitation learning},
    author={Jiang, Zhenyu and Xie, Yuqi and Lin, Kevin and Xu, Zhenjia and Wan, Weikang and Mandlekar, Ajay and Fan, Linxi and Zhu, Yuke},
    journal={arXiv preprint arXiv:2410.24185},
    year={2024}
}

@inproceedings{
    garrett2024skillgen,
    title={SkillGen: Automated Demonstration Generation for Efficient Skill Learning and Deployment},
    author={Caelan Reed Garrett and Ajay Mandlekar and Bowen Wen and Dieter Fox},
    booktitle={8th Annual Conference on Robot Learning},
    year={2024},
    url={https://openreview.net/forum?id=YOFrRTDC6d}
}

@article{zhou2025reinforcegen,
    title={ReinforceGen: Hybrid Skill Policies with Automated Data Generation and Reinforcement Learning},
    author={Zhou, Zihan and Garg, Animesh and Mandlekar, Ajay and Garrett, Caelan},
    journal={arXiv preprint arXiv:2512.16861},
    year={2025}
}

@inproceedings{
    lin2025cpgen,
    title={Constraint-Preserving Data Generation for Visuomotor Policy Generalization},
    author={Kevin Lin and Varun Ragunath and Andrew McAlinden and Aaditya Prasad and Jimmy Wu and Yuke Zhu and Jeannette Bohg},
    booktitle={9th Annual Conference on Robot Learning},
    year={2025},
    url={https://openreview.net/forum?id=KSKzA1mwKs}
}

@article{li2025momagen,
  title={Momagen: Generating demonstrations under soft and hard constraints for multi-step bimanual mobile manipulation},
  author={Li, Chengshu and Xu, Mengdi and Bahety, Arpit and Yin, Hang and Jiang, Yunfan and Huang, Huang and Wong, Josiah and Garlanka, Sujay and Gokmen, Cem and Zhang, Ruohan and others},
  journal={arXiv preprint arXiv:2510.18316},
  year={2025}
}

@article{yang2025physics,
    title={Physics-driven data generation for contact-rich manipulation via trajectory optimization},
    author={Yang, Lujie and Suh, HJ and Zhao, Tong and Graesdal, Bernhard Paus and Kelestemur, Tarik and Wang, Jiuguang and Pang, Tao and Tedrake, Russ},
    journal={arXiv preprint arXiv:2502.20382},
    year={2025}
}

@article{liu2025manipulation,
    title={Manipulation as in simulation: Enabling accurate geometry perception in robots},
    author={Liu, Minghuan and Zhu, Zhengbang and Han, Xiaoshen and Hu, Peng and Lin, Haotong and Li, Xinyao and Chen, Jingxiao and Xu, Jiafeng and Yang, Yichu and Lin, Yunfeng and others},
    journal={arXiv preprint arXiv:2509.02530},
    year={2025}
}

@INPROCEEDINGS{curobo_icra23,
    author={Sundaralingam, Balakumar and Hari, Siva Kumar Sastry and
            Fishman, Adam and Garrett, Caelan and Van Wyk, Karl and Blukis, Valts and
            Millane, Alexander and Oleynikova, Helen and Handa, Ankur and
            Ramos, Fabio and Ratliff, Nathan and Fox, Dieter},
    booktitle={2023 IEEE International Conference on Robotics and Automation (ICRA)},
    title={CuRobo: Parallelized Collision-Free Robot Motion Generation},
    year={2023},
    volume={},
    number={},
    pages={8112-8119},
    doi={10.1109/ICRA48891.2023.10160765}
}

@misc{curobo_report23,
    title={cuRobo: Parallelized Collision-Free Minimum-Jerk Robot Motion Generation},
    author={Balakumar Sundaralingam and Siva Kumar Sastry Hari and Adam Fishman and Caelan Garrett
          and Karl Van Wyk and Valts Blukis and Alexander Millane and Helen Oleynikova and Ankur Handa
          and Fabio Ramos and Nathan Ratliff and Dieter Fox},
    year={2023},
    eprint={2310.17274},
    archivePrefix={arXiv},
    primaryClass={cs.RO}
 }

@article{inui2016shrinking,
    title={Shrinking sphere: A parallel algorithm for computing the thickness of 3D objects},
    author={Inui, Masatomo and Umezu, Nobuyuki and Shimane, Ryohei},
    journal={Computer-Aided Design and Applications},
    volume={13},
    number={2},
    pages={199--207},
    year={2016},
    publisher={Taylor \& Francis}
}

@article{khatib1987unified,
    title={A unified approach for motion and force control of robot manipulators: The operational space formulation},
    author={Khatib, Oussama},
    journal={IEEE Journal on Robotics and Automation},
    volume={3},
    number={1},
    pages={43--53},
    year={1987},
    publisher={IEEE}
}

@inproceedings{pomerleau1989alvinn,
    title={Alvinn: An autonomous land vehicle in a neural network},
    author={Pomerleau, Dean A},
    booktitle={Advances in neural information processing systems},
    pages={305--313},
    year={1989}
}

@inproceedings{bain1995framework,
    title={A Framework for Behavioural Cloning.},
    author={Bain, Michael and Sammut, Claude},
    booktitle={Machine intelligence 15},
    pages={103--129},
    year={1995}
}

@inproceedings{todorov2012mujoco,
    title={Mujoco: A physics engine for model-based control},
    author={Todorov, Emanuel and Erez, Tom and Tassa, Yuval},
    booktitle={IEEE/RSJ International Conference on Intelligent Robots and Systems},
    pages={5026--5033},
    year={2012},
}

@inproceedings{robosuite2020,
    title={robosuite: A Modular Simulation Framework and Benchmark for Robot Learning},
    author={Yuke Zhu and Josiah Wong and Ajay Mandlekar and Roberto Mart{\'i}n-Mart{\'i}n},
    booktitle={arXiv preprint arXiv:2009.12293},
    year={2020}
}

@inproceedings{robomimic2021,
    title={What Matters in Learning from Offline Human Demonstrations for Robot Manipulation},
    author={Ajay Mandlekar and Danfei Xu and Josiah Wong and Soroush Nasiriany and Chen Wang and Rohun Kulkarni and Li Fei-Fei and Silvio Savarese and Yuke Zhu and Roberto Mart\'{i}n-Mart\'{i}n},
    booktitle={Conference on Robot Learning (CoRL)},
    year={2021}
}

@article{maddukuri2025sim,
    title={Sim-and-real co-training: A simple recipe for vision-based robotic manipulation},
    author={Maddukuri, Abhiram and Jiang, Zhenyu and Chen, Lawrence Yunliang and Nasiriany, Soroush and Xie, Yuqi and Fang, Yu and Huang, Wenqi and Wang, Zu and Xu, Zhenjia and Chernyadev, Nikita and others},
    journal={arXiv preprint arXiv:2503.24361},
    year={2025}
}

@article{black2024pi_0,
  title={$\pi_0 $: A Vision-Language-Action Flow Model for General Robot Control},
  author={Black, Kevin and Brown, Noah and Driess, Danny and Esmail, Adnan and Equi, Michael and Finn, Chelsea and Fusai, Niccolo and Groom, Lachy and Hausman, Karol and Ichter, Brian and others},
  journal={arXiv preprint arXiv:2410.24164},
  year={2024}
}

@article{bjorck2025gr00t,
    title={Gr00t n1: An open foundation model for generalist humanoid robots},
    author={Bjorck, Johan and Casta{\~n}eda, Fernando and Cherniadev, Nikita and Da, Xingye and Ding, Runyu and Fan, Linxi and Fang, Yu and Fox, Dieter and Hu, Fengyuan and Huang, Spencer and others},
    journal={arXiv preprint arXiv:2503.14734},
    year={2025}
}

@article{hu2024adaflow,
  title={AdaFlow: Imitation Learning with Variance-Adaptive Flow-Based Policies},
  author={Hu, Xixi and Liu, Bo and Liu, Xingchao and Liu, Qiang},
  journal={arXiv preprint arXiv:2402.04292},
  year={2024}
}

@article{gu2025humanoid,
    title={Humanoid locomotion and manipulation: Current progress and challenges in control, planning, and learning},
    author={Gu, Zhaoyuan and Li, Junheng and Shen, Wenlan and Yu, Wenhao and Xie, Zhaoming and McCrory, Stephen and Cheng, Xianyi and Shamsah, Abdulaziz and Griffin, Robert and Liu, C Karen and others},
    journal={arXiv preprint arXiv:2501.02116},
    year={2025}
}

@misc{sferrazza2024humanoidbench,
    title={HumanoidBench: Simulated Humanoid Benchmark for Whole-Body Locomotion and Manipulation},
    author={Carmelo Sferrazza and Dun-Ming Huang and Xingyu Lin and Youngwoon Lee and Pieter Abbeel},
    year={2024},
}

@article{luo2024smplolympics,
  title={Smplolympics: Sports environments for physically simulated humanoids},
  author={Luo, Zhengyi and Wang, Jiashun and Liu, Kangni and Zhang, Haotian and Tessler, Chen and Wang, Jingbo and Yuan, Ye and Cao, Jinkun and Lin, Zihui and Wang, Fengyi and others},
  journal={arXiv preprint arXiv:2407.00187},
  year={2024}
}

@inproceedings{he2025hover,
    title={Hover: Versatile neural whole-body controller for humanoid robots},
    author={He, Tairan and Xiao, Wenli and Lin, Toru and Luo, Zhengyi and Xu, Zhenjia and Jiang, Zhenyu and Kautz, Jan and Liu, Changliu and Shi, Guanya and Wang, Xiaolong and others},
    booktitle={2025 IEEE International Conference on Robotics and Automation (ICRA)},
    pages={9989--9996},
    year={2025},
    organization={IEEE}
}

@article{luo2025sonic,
    title={Sonic: Supersizing motion tracking for natural humanoid whole-body control},
    author={Luo, Zhengyi and Yuan, Ye and Wang, Tingwu and Li, Chenran and Chen, Sirui and Casta{\~n}eda, Fernando and Cao, Zi-Ang and Li, Jiefeng and Minor, David and Ben, Qingwei and others},
    journal={arXiv preprint arXiv:2511.07820},
    year={2025}
}

@article{ben2025homie,
  title={Homie: Humanoid loco-manipulation with isomorphic exoskeleton cockpit},
  author={Ben, Qingwei and Jia, Feiyu and Zeng, Jia and Dong, Junting and Lin, Dahua and Pang, Jiangmiao},
  journal={arXiv preprint arXiv:2502.13013},
  year={2025}
}

@article{brohan2022rt,
    title={Rt-1: Robotics transformer for real-world control at scale},
    author={Brohan, Anthony and Brown, Noah and Carbajal, Justice and Chebotar, Yevgen and Dabis, Joseph and Finn, Chelsea and Gopalakrishnan, Keerthana and Hausman, Karol and Herzog, Alex and Hsu, Jasmine and others},
    journal={arXiv preprint arXiv:2212.06817},
    year={2022}
}

@inproceedings{o2024open,
    title={Open x-embodiment: Robotic learning datasets and rt-x models: Open x-embodiment collaboration 0},
    author={O’Neill, Abby and Rehman, Abdul and Maddukuri, Abhiram and Gupta, Abhishek and Padalkar, Abhishek and Lee, Abraham and Pooley, Acorn and Gupta, Agrim and Mandlekar, Ajay and Jain, Ajinkya and others},
    booktitle={2024 IEEE Int'l Conf on Robotics and Automation (ICRA)},
    ign-pages={6892--6903},
    year={2024},
    ign-organization={IEEE}
}

@INPROCEEDINGS{ebert2021bridge, 
    AUTHOR    = {Frederik Ebert AND Yanlai Yang AND Karl Schmeckpeper AND Bernadette Bucher AND Georgios Georgakis AND Kostas Daniilidis AND Chelsea Finn AND Sergey Levine}, 
    TITLE     = {{Bridge Data: Boosting Generalization of Robotic Skills with Cross-Domain Datasets}}, 
    BOOKTITLE = {Robotics: Science and Systems}, 
    YEAR      = {2022}, 
    ign-month     = {June}, 
    ign-doi       = {10.15607/RSS.2022.XVIII.063} 
}

@article{khazatsky2024droid,
  title={Droid: A large-scale in-the-wild robot manipulation dataset},
  author={Khazatsky, Alexander and Pertsch, Karl and Nair, Suraj and Balakrishna, Ashwin and Dasari, Sudeep and Karamcheti, Siddharth and Nasiriany, Soroush and Srirama, Mohan Kumar and Chen, Lawrence Yunliang and Ellis, Kirsty and others},
  journal={arXiv preprint arXiv:2403.12945},
  year={2024}
}

@article{wei2025empirical,
  title={Empirical analysis of sim-and-real cotraining of diffusion policies for planar pushing from pixels},
  author={Wei, Adam and Agarwal, Abhinav and Chen, Boyuan and Bosworth, Rohan and Pfaff, Nicholas and Tedrake, Russ},
  journal={arXiv preprint arXiv:2503.22634},
  year={2025}
}

@article{cheng2025generalizable,
  title={Generalizable domain adaptation for sim-and-real policy co-training},
  author={Cheng, Shuo and Ma, Liqian and Chen, Zhenyang and Mandlekar, Ajay and Garrett, Caelan and Xu, Danfei},
  journal={arXiv preprint arXiv:2509.18631},
  year={2025}
}

@article{haldar2026point,
  title={Point Bridge: 3D Representations for Cross Domain Policy Learning},
  author={Haldar, Siddhant and Johannsmeier, Lars and Pinto, Lerrel and Gupta, Abhishek and Fox, Dieter and Narang, Yashraj and Mandlekar, Ajay},
  journal={arXiv preprint arXiv:2601.16212},
  year={2026}
}

@article{saxena2025matters,
  title={What matters in learning from large-scale datasets for robot manipulation},
  author={Saxena, Vaibhav and Bronars, Matthew and Arachchige, Nadun Ranawaka and Wang, Kuancheng and Shin, Woo Chul and Nasiriany, Soroush and Mandlekar, Ajay and Xu, Danfei},
  journal={arXiv preprint arXiv:2506.13536},
  year={2025}
}

@article{barreiros2025careful,
  title={A careful examination of large behavior models for multitask dexterous manipulation},
  author={Barreiros, Jose and Beaulieu, Andrew and Bhat, Aditya and Cory, Rick and Cousineau, Eric and Dai, Hongkai and Fang, Ching-Hsin and Hashimoto, Kunimatsu and Irshad, Muhammad Zubair and Itkina, Masha and others},
  journal={arXiv preprint arXiv:2507.05331},
  year={2025}
}

@article{schaal1999imitation,
  title={Is imitation learning the route to humanoid robots?},
  author={Schaal, Stefan},
  journal={Trends in cognitive sciences},
  volume={3},
  ign-number={6},
  ign-pages={233--242},
  year={1999},
  ign-publisher={Elsevier}
}

@article{Ijspeert2002MovementIW,
  title={Movement imitation with nonlinear dynamical systems in humanoid robots},
  author={Auke Jan Ijspeert and Jun Nakanishi and Stefan Schaal},
  journal={Proceedings 2002 IEEE Int'l Conf on Robotics and Automation},
  year={2002},
  volume={2},
  ign-pages={1398-1403 vol.2}
}

@inproceedings{Billard2008RobotPB,
  title={Robot Programming by Demonstration},
  author={Aude Billard and Sylvain Calinon and R{\"u}diger Dillmann and Stefan Schaal},
  booktitle={Springer Handbook of Robotics},
  year={2008}
}

@article{Calinon2010LearningAR,
  title={Learning and Reproduction of Gestures by Imitation},
  author={Sylvain Calinon and Florent D'halluin and Eric L. Sauser and Darwin G. Caldwell and Aude Billard},
  journal={IEEE Robotics and Automation Magazine},
  year={2010},
  volume={17},
  ign-pages={44-54}
}

@article{chi2023diffusion,
  title={Diffusion policy: Visuomotor policy learning via action diffusion},
  author={Chi, Cheng and Xu, Zhenjia and Feng, Siyuan and Cousineau, Eric and Du, Yilun and Burchfiel, Benjamin and Tedrake, Russ and Song, Shuran},
  journal={The Int'l Journal of Robotics Research},
  ign-pages={02783649241273668},
  year={2023},
  ign-publisher={SAGE Publications Sage UK: London, England}
}

@article{brohan2023rt,
  title={Rt-2: Vision-language-action models transfer web knowledge to robotic control},
  author={Brohan, Anthony and Brown, Noah and Carbajal, Justice and Chebotar, Yevgen and Chen, Xi and Choromanski, Krzysztof and Ding, Tianli and Driess, Danny and Dubey, Avinava and Finn, Chelsea and others},
  journal={arXiv preprint arXiv:2307.15818},
  year={2023}
}

@article{tian2025interndata,
  title={InternData-A1: Pioneering High-Fidelity Synthetic Data for Pre-training Generalist Policy},
  author={Tian, Yang and Yang, Yuyin and Xie, Yiman and Cai, Zetao and Shi, Xu and Gao, Ning and Liu, Hangxu and Jiang, Xuekun and Qiu, Zherui and Yuan, Feng and others},
  journal={arXiv preprint arXiv:2511.16651},
  year={2025}
}

@article{yin2026genie,
  title={Genie Sim 3.0: A High-Fidelity Comprehensive Simulation Platform for Humanoid Robot},
  author={Yin, Chenghao and Huang, Da and Yang, Di and Wang, Jichao and Zhao, Nanshu and Xu, Chen and Sun, Wenjun and Hou, Linjie and Li, Zhijun and Wu, Junhui and others},
  journal={arXiv preprint arXiv:2601.02078},
  year={2026}
}

@article{yang2025omniretarget,
  title={Omniretarget: Interaction-preserving data generation for humanoid whole-body loco-manipulation and scene interaction},
  author={Yang, Lujie and Huang, Xiaoyu and Wu, Zhen and Kanazawa, Angjoo and Abbeel, Pieter and Sferrazza, Carmelo and Liu, C Karen and Duan, Rocky and Shi, Guanya},
  journal={arXiv preprint arXiv:2509.26633},
  year={2025}
}

\appendix
\newpage

\section*{Overview}

The appendix contains the following content.

\begin{itemize}

    \item \textbf{\hyperref[app:skill]{Skill Planning Example}} (Appendix~\ref{app:skill}): example of skill planning on the Table-To-Shelf task.

    \item \textbf{\hyperref[app:pseudocode]{\sysName{} Pseudocode}} (Appendix~\ref{app:pseudocode}): system-level pseudocode for \sysName{} and whole-body skill adaptation.

    \item \textbf{\hyperref[app:manipulation]{Motion Planning Details}} (Appendix~\ref{app:manipulation}): collision representation and whole-body inverse kinematics details.

    \item \textbf{\hyperref[app:benchmark]{Loco-Manipulation Benchmark}} (Appendix~\ref{app:benchmark}): description of the humanoid loco-manipulation simulation benchmark and tasks.

    \item \textbf{\hyperref[app:dexmimicgen-plus]{DexMimicGen+ Baseline}} (Appendix~\ref{app:dexmimicgen-plus}): implementation details for the DexMimicGen+ baseline.

    \item \textbf{\hyperref[app:g1_vs_g1floating]{Scaling Across Embodiments}} (Appendix~\ref{app:g1_vs_g1floating}): comparison between legged and floating-base humanoid embodiments.
    
    \item \textbf{\hyperref[app:policy-training]{Policy Training Details}} (Appendix~\ref{app:policy-training}): VLA training and evaluation details.

    \item \textbf{\hyperref[app:raw-tables]{Raw Tables for Bar Plots}} (Appendix~\ref{app:raw-tables}): raw tables corresponding to ablation plots.

    \item \textbf{\hyperref[app:real-world-scoring]{Real-World Experiment Details}} (Appendix~\ref{app:real-world}): Details on hardware and scoring criteria for real-world experiments.
\end{itemize}

\vspace{2mm}

\section{Skill Planning Example}\label{app:skill}

\begin{figure*}[ht]
    \centering
    \includegraphics[width=\linewidth]{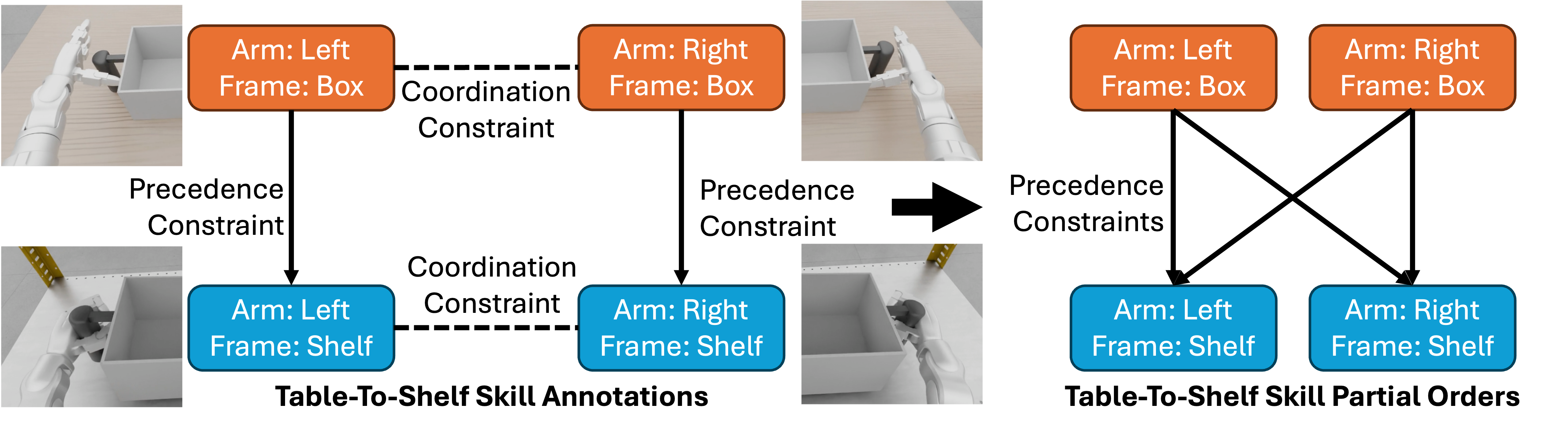}
    \caption{\textbf{Precedence and Coordination Constraints.} Skill planning visualized on the Table-to-Shelf task.
    A human annotates precedence and coordination constraints among the skills ({\it left}). \sysName{} automatically compiles these constraints into partial orders on the skills, defining legal execution orders ({\it right}).}
    \label{fig:orders}
\end{figure*}

Continuing on from Section~\ref{sec:skill}, Fig.~\ref{fig:orders} visualizes the input constraint annotation ({\it left}) as well as the output Directed Acyclic Graph (DAG) that \sysName{} produces ({\it right}) for the Table-to-Shelf task (Fig.~\ref{fig:teaser}). 
\proc{\sysName{}} executes this in two iterations. 
First, both pick skills are executed concurrently. 
Then, both place skills are also executed concurrently.
These iterations correspond to the two layers in the DAG when traversed via a topological sort.

\section{\sysName{} Pseudocode}\label{app:pseudocode}

Algorithms~\ref{alg:system} and \ref{alg:adaptation} give the pseudocode for \proc{\sysName{}} and \proc{adapt-skill-demos} respectively, the core \sysName{} algorithms in Section~\ref{sec:method}.

\begin{algorithm} %
\begin{footnotesize} %
  \caption{\sysName{} Pseudocode} 
  \label{alg:system}  
  \begin{algorithmic}[1] %
    \Procedure{\sysName{}}{$s_0, \langle \Psi, {\cal P} \rangle$}
    \State $s \gets \kw{copy}(s_0)$ \Comment{Current state}
    \While{$|\Psi| \neq 0$}
        \State $\Psi_i \gets \{\psi \in \Psi \mid \neg \exists \psi' \in \Psi. \langle \psi', \psi \rangle \in {\cal P}\}$
        \State $\Psi \gets \Psi \setminus \Psi_i$ \Comment{Subtract current skills $\Psi_i$ from $\Psi$}
        \State $T \gets \{\;\}$ \Comment{End-effector pose targets}
        \For{$\langle e, f, d^\psi \rangle \in \Psi_i$} \Comment{Skill $\psi$ for end-effector $e$}
            \State $\langle s_0^\psi, \_, \_ \rangle \gets d^\psi[0]$ \Comment{Reference state $s_0^\psi$}
            \State $T[e] \gets s[f] \inv{s^\psi_0[f]} s_0^\psi[e]$ \Comment{Target pose for $e$}
        \EndFor
        \State $q \gets s[{\cal J}]$ \Comment{Current configuration}
        \State $Q \gets \proc{whole-inv-kinematics}({\cal J}, T, s)$
        \For{$q'' \in Q$} \Comment{IK configurations batch}
            \State $q' \gets \kw{copy}(q)$ \Comment{Target switch configuration}
            \State $q'[J_l] \gets q''[J_l]$ \Comment{Set the locomotion joints}
            \State $\tau_l \gets \proc{plan-motion}(J_l, q, q', s)$ \Comment{Legs} %
            \If{$\tau_l \neq \kw{None}$} %
                \State $s \gets \proc{control-locomotion}(\tau_l)$
                \State \kw{break} \Comment{Planning success}
            \EndIf
        \EndFor
        \State \kw{else}
        \State \;\;\;\;\;\Return \kw{False} \Comment{Planning failure}
        \State $q' \gets s[{\cal J}]$ \Comment{Achieved switch configuration}
        \State $\tau_m \gets \proc{plan-motion}(J_{t} \cup J_{a_l} \cup J_{a_r}, q', q'', s)$ \Comment{Arms} %
        \State $s \gets \proc{control-manipulation}(\tau_m)$
        \State $\tau_{\Psi_i} \gets \proc{adapt-skill-demos}(s, \Psi_i)$
        \State $s \gets \proc{control-manipulation}(\tau_{\Psi_i})$
    \EndWhile
    \State \Return $\proc{check-success}(s)$ \Comment{Execution success}
    \EndProcedure
\end{algorithmic}
\end{footnotesize}
\end{algorithm}

\section{Motion Planning Details}\label{app:manipulation}

We adapt skills to new environments and states by solving for new robot configurations that satisfy the same constraints as the demonstrations, while responding to changes in pose and avoiding collisions.
We develop our algorithms on top of cuRobo~\cite{curobo_icra23,curobo_report23} to take advantage of GPU-accelerated collision checking and batch inverse kinematics.

\begin{algorithm} %
\begin{footnotesize} %
  \caption{Whole-Body Skill Adaptation Pseudocode} 
  \label{alg:adaptation}  
  \begin{algorithmic}[1] %
    \Procedure{adapt-skill-demos}{$s, \Psi_i$}
    \State $\tau \gets [\;]$ \Comment{Adapted skill trajectory}
        \While{\kw{True}}
            \State $T \gets \{\;\}$ \Comment{End-effector pose targets}
            \For{$\langle e, f, d^\psi \rangle \in \Psi_i$} \Comment{Skill $\psi$ for end-effector $e$}
                \State $\langle s_0^\psi, \_, \_ \rangle \gets d^\psi[0]$ \Comment{Reference state $s_0^\psi$}
                \If{$|\tau| \leq |d^\psi| - 1$}
                \State $\langle \_, \_, a^\psi \rangle \gets d^\psi[|\tau|]$ \Comment{Skill demo action $a^\psi$}
                \State $T[e] \gets s[f] \inv{s^\psi_0[f]} a^\psi[e]$ \Comment{Adapt $a$ to $s$}
                \EndIf
            \EndFor
            \If{$|T| = 0$} \Comment{All skill actions processed}
                \State \Return $\tau_{\Psi_i}$
            \EndIf
            \State $q' \gets \proc{whole-inv-kinematics}(J_{t} \cup J_{a_l} \cup J_{a_r}, T, s)[0]$
            \State $\tau \gets \tau + [q']$ \Comment{Add the next waypoint $q$}
        \EndWhile
    \EndProcedure
\end{algorithmic}
\end{footnotesize}
\end{algorithm}

\begin{figure}[h]
    \centering
    \includegraphics[width=\columnwidth]{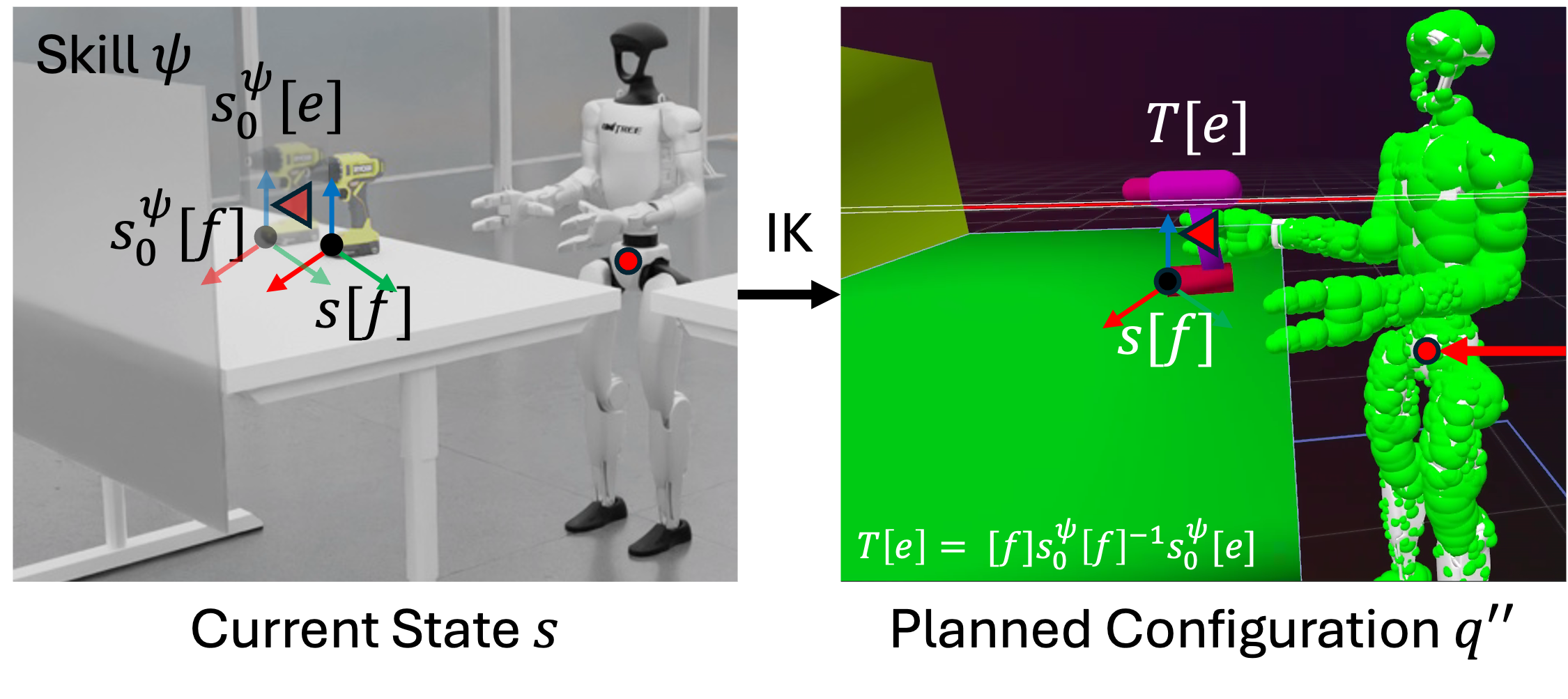}
    \caption{\textbf{Whole-Body Skill Adaptation.} {\it Left:} skill $\psi$ for end-effector $e$ relative to object frame $f$ is adapted to a new state $s$. {\it Right}: spherical collision representation used for planning and IK configuration $q''$ for adapted skill target pose $T[e]$.
    }
    \label{fig:adaptation}
\end{figure}

\textbf{Humanoid Collision Representation.}
To leverage GPU-accelerated collision checking, we represent the robot and all moving objects as a set of spheres with varying radii.
Instead of manually placing these spheres per robot~\cite{curobo_icra23,curobo_report23}, which is challenging for humanoids due to, for example, the intricacy of their dexterous hands, we take an automated approach.
For each mesh on the robot or object, we sample a large set of points on its surface.
For each sampled point, we create a candidate sphere by solving for the minimum volume sphere that is tangent to the sampled point and another face on the mesh~\cite{inui2016shrinking}.
We then inflate the sphere radii by a hyperparameter $\epsilon \approx 0.01m$ to cover more of the mesh's surface.
Finally, we use a greedy combinatorial optimization algorithm to select a set of $N$ spheres that maximizes coverage of the mesh's surface subject to a computation budget.
Figure~\ref{fig:adaptation} ({\it right}) displays a set of collision spheres automatically computed for the G1 robot. 
To model collisions with held objects, we detect which movable objects are in contact with a robot end-effector then lazily compute a sphere decomposition for them.

Representing the robot as a set of spheres over-approximates its geometry.
Because we consider inverse kinematics and motion planning subproblems induced through manipulation, the initial configuration and goal end-effector poses are often in contact with manipulable objects, which are collision states causing the subproblem to be infeasible, especially due to sphere over-approximation.
However, intuitively, the robot can safely make contact between certain pairs of robot and object bodies.
To account for this, we compute the spheres currently in collision before each IK or planning call and shrink them until they are no longer in collision.
We do the same for target end-effector poses, but only for the robot links in the same rigid (given the planning joints) connected component of the kinematic tree as the end-effector.

\textbf{Whole-Body Inverse Kinematics.}
We leverage cuRobo to implement \proc{whole-inv-kinematics}, which performs multi-link batch inverse kinematics.
Sometimes, the robot can reach the start of the next skill without moving its legs.
In such cases, it is preferable to skip locomotion planning, saving computation time, avoiding the use of both control modalities, and reducing the risk of control errors during dynamic walking.
Additionally, because torso motion affects both end-effectors, it is desirable to minimize unneeded torso movement away from the default torso posture.
To do this, we minimize the $L_0$ distance from the current configuration $||q'' - q||_0$, weighted by joint group.
We achieve this by performing an iterative optimization, where we lock a subset of the joints $J \subseteq {\cal J}$ to have their current positions $q[J]$.
In our experiments, we considered the free joint order $[J_a, J_a \cup J_t, J_a \cup J_l, {\cal J}]$, which corresponds to allowing 1) arm motion, 2) arm and torso motion, 3) arm and leg motion, and 4) free motion.

\section{Loco-Manipulation Benchmark} %
\label{app:benchmark}

We developed a humanoid \textbf{loco-manipulation benchmark} (Figure \ref{fig:g1_benchmark}) built on robosuite~\cite{robosuite2020} and MuJoCo~\cite{todorov2012mujoco}. 
This benchmark specifically tests loco-manipulation; success critically depends on accurate base motion and whole-body coordination. 
This benchmark enables controlled comparisons of data generation strategies, policy architectures, and embodiment choices in loco-manipulation settings, as analyzed in Section~\ref{sec:experiments}.

The benchmark contains \textbf{9 tasks} involving a simulated G1 humanoid robot.
Tasks vary along three axes: (i) the extent of required base motion (from minimal repositioning along one axis to multi-stage navigation), (ii) object interaction complexity (single-arm, bimanual, and whole-body), and (iii) execution horizon. 
Each task specifies a goal represented as a binary success condition.
Initial states are generated by randomly sampling object poses and the robot's root pose.
Together, these tasks emphasize the locomotion--manipulation interface: small errors in foot placement directly impact reachability.

\begin{footnotesize}
\begin{enumerate}
    \item \textbf{Box Lift Floor}: Grasp the box from the floor and lift to a target height.
    
    \item \textbf{Push Button}: Approach an industrial panel and press its button.
    
    \item \textbf{Box Lift}: Approach the table, grasp the box, and lift to a target height.
    
    \item \textbf{Push Shelf Forward}: Push the shelving cart into a marked target zone.
    
    \item \textbf{Drill Lift}: Approach table, grasp drill, and lift to target height.
    
    \item \textbf{Drill PnP}: Pick a drill from one table and place it on a second table.
    
    \item \textbf{Box Table To Shelf}: Transfer a box from a table into a shelf.
    
    \item \textbf{Pick Drill From Holder}: Extract a drill from a holder and lift it to a target height.
    
    \item \textbf{Drill Lift Obstacle}: Navigate around a blocking shelf to reach the table, then grasp and lift the drill to a target height.
\end{enumerate}
\end{footnotesize}

In Table~\ref{tab:benchmark_capabilities}, we further describe the tasks in the G1 Loco-manipulation Benchmark according to the attributes they contain.

\begin{table}[h]
\centering
\caption{Capabilities tested by each benchmark task.}
\label{tab:benchmark_capabilities}

\resizebox{\columnwidth}{!}{%
\begin{tabular}{l
  c c c c c c c
}
\toprule
Task
& {Loco}
& {Nav}
& {1-arm}
& {2-arm}
& {Vert}
& {Contact}
& {Long} \\
\midrule
PushButton          & \cmark &        & \cmark &        &        &        &        \\
DrillLift           & \cmark &        & \cmark &        &        &        &        \\
BoxLift             & \cmark &        &        & \cmark &        &        &        \\
BoxLiftFloor        & \cmark &        &        & \cmark & \cmark &        &        \\
PickDrillFromHolder & \cmark &        & \cmark &        &        & \cmark &        \\
PushShelfForward    & \cmark &        &        & \cmark &        & \cmark &        \\
DrillLiftObstacle   & \cmark & \cmark & \cmark &        &        &        & \cmark \\
DrillPnP            & \cmark &        & \cmark &        &        &        & \cmark \\
BoxTableToShelf     & \cmark &        &        & \cmark & \cmark &        & \cmark \\
\bottomrule
\end{tabular}%
}

\vspace{2pt}
\footnotesize
Capabilities: \textbf{Loco} = locomotion; \textbf{Nav} = obstacle-free navigation;
\textbf{1-arm} = single-arm manipulation; \textbf{2-arm} = bimanual coordination;
\textbf{Vert} = vertical reach; \textbf{Contact} = contact-rich interaction;
\textbf{Long} = long-horizon execution.
\end{table}

\section{DexMimicGen+ Baseline}\label{app:dexmimicgen-plus}

As described in Section~\ref{subsec:features}, we compare \sysName{} against a DexMimicGen+ baseline in order to highlight the importance of intelligent whole-body planning and adaptation.
The DexMimicGen+ baseline extends DexMimicGen~\cite{jiang2024dexmimicgen} with the ability to handle locomotion when needed, as the original work solely focused on stationary bimanual manipulation. 
DexMimicGen+ inherits the same action space that we propose for \sysName{} (Section~\ref{sec:control}), where locomotion actions are represented as $x, y, \theta$ velocity commands along with an additional $z$ height command.
No motion planning is used for this baseline, to stay consistent with the original work.
We make the following modifications to the DexMimicGen algorithm:
\begin{enumerate}
    \item We manually annotate source demonstration subtasks that will require locomotion.
    \item For each subtask that requires locomotion, we invoke \proc{whole-inv-kinematics} to infer a target base pose for one (or both) arm poses at the start of each subtask. The \proc{whole-inv-kinematics} procedure does not consider collisions, and all joints apart from the legs are unlocked and free to move from the current configuration.
    \item To move from a current base configuration to a new base configuration, a straight-line interpolated path is used (similar to interpolation segments for the arms in DexMimicGen~\cite{jiang2024dexmimicgen}).
\end{enumerate}
This baseline lacks several crucial features introduced by \sysName{}, including the use of skill reasoning, motion planning for locomotion and arm movement, and collision checking.

\section{Scaling Across Embodiments}
\label{app:g1_vs_g1floating}

While our main results focus on the G1 humanoid, \sysName{} also applies to alternative embodiments with different locomotion and manipulation constraints. 
We demonstrate data generation on a floating-base humanoid variant using the same pipeline without algorithmic changes and present policy success rates on generated data in Table~\ref{tab:embodiment}.
We find that policies trained with and evaluated on the floating body humanoid perform similarly on average to those in the legged setting.

\begin{table}[h]
\centering

\begin{tabular}{l
  c
  c
}
\toprule
Task 
& {G1}
& {G1 w/o Legs} \\
\midrule
Box Lift Floor         & \textbf{0.97} & \textbf{1.00} \\
Push Button            & 0.92 & \textbf{1.00} \\
Box Lift               & \textbf{1.00} & \textbf{1.00} \\
Push Shelf Forward     & \textbf{1.00} & 0.53 \\
Drill Lift             & \textbf{1.00} & \textbf{1.00} \\
Drill PnP              & \textbf{0.70} & 0.57 \\
Box Table to Shelf     & 0.53 & \textbf{1.00} \\
Pick Drill from Holder & \textbf{1.00} & \textbf{1.00} \\
Drill Lift Obstacle    & 0.87 & \textbf{1.00} \\
\midrule
Average                & 0.89 & \textbf{0.90} \\
\bottomrule
\end{tabular}
\vspace{5mm}
\caption{Embodiment comparison under the \sysName{} pipeline. Policy success rate (PSR). Floating-body variants perform similarly on average to legged humanoid variants.
}
\label{tab:embodiment}
\end{table}

\section{Policy Training Details}\label{app:policy-training}

We additionally include details from post-training a vision-language action model.

\textbf{Training Configurations.}
The VLA model in our policy architecture ablation table (Table \ref{tab:policy_ablations_abs}) is initialized from the pre-trained GR00T N1.6 base model \cite{bjorck2025gr00t}. For post-training of the VLA, we use a learning rate of 1e-5  with a global batch size of 128 and train for 25,000 optimization steps. The policy uses a single egocentric camera as visual input with an image resolution of 224x224. Training is performed on a single node equipped with eight NVIDIA H100 GPUs.

\textbf{Evaluation Details.}
During evaluation, each task is executed for 100 rollout trials to obtain statistically reliable performance metrics. Model checkpoints are saved every 5,000 training steps, and evaluation is performed across all saved checkpoints. The final reported results correspond to the checkpoint that achieves the best evaluation performance. Both training and evaluation are conducted on the same hardware platform to ensure consistency across experimental settings.

\section{Raw Tables for Bar Plots}\label{app:raw-tables}

We additionally provide raw tables for the bar plots for policy architecture ablations in Table \ref{tab:policy_ablations_abs} and data generation ablations in Table \ref{tab:g1_noise_ablation_psr}.

\begin{table}[H]
\centering
\begin{tabular}{l
  c
  c
  c
}
\toprule
Task &
{VLA} &
{DP} &
{Flow Matching} \\
\midrule
Box Lift Floor          & \bfseries 0.97 & 0.92 & \bfseries 1.00 \\
Push Button             & 0.92 & 0.55 & \bfseries 1.00 \\
Box Lift                & \bfseries 1.00 & 0.95 & 0.98 \\
Push Shelf Forward      & \bfseries 1.00 & 0.95 & 0.93 \\
Drill Lift              & \bfseries 1.00 & 0.12 & \bfseries 1.00 \\
Drill PnP               & \bfseries 0.70 & 0.10 & 0.50 \\
Box Table to Shelf      & \bfseries 0.53 & 0.02 & 0.52 \\
Pick Drill From Holder  & \bfseries 1.00 & 0.95 & 0.96 \\
Drill Lift Obstacle     & 0.87 & 0.06 & \bfseries 0.88 \\
\midrule
Average                 & \bfseries \textbf{0.89} & 0.51 & 0.86 \\
\bottomrule
\end{tabular}
\vspace{5mm}
\caption{\textbf{Policy architecture ablation.}
All policies are trained on 1,000 demonstrations generated by \sysName{} and evaluated using the best-performing checkpoint. VLA achieves the highest average success rate, while Flow Matching performs competitively across several tasks.}
\label{tab:policy_ablations_abs}
\end{table}

\begin{table}[H]
\centering
\begin{tabular}{l
  c
  c
  c
}
\toprule
{Task}
& {Ours}
& {w/o Motion Noise}
& {w/o Init. Noise} \\
\midrule
Box Lift Floor          & 0.97 & \textbf{1.00} & \textbf{1.00} \\
Push Button            & 0.92 & \textbf{1.00} & 0.43 \\
Box Lift               & \textbf{1.00} & 0.90 & \textbf{1.00} \\
Push Shelf Forward     & \textbf{1.00} & 0.70 & 0.93 \\
Drill Lift             & \textbf{1.00} & 0.23 & 0.23 \\
Drill PnP              & \textbf{0.70} & 0.17 & 0.10 \\
Box Table to Shelf     & \textbf{0.53} & 0.00 & 0.37 \\
Pick Drill from Holder & \textbf{1.00} & 0.26 & 0.35 \\
Drill Lift Obstacle    & \textbf{0.87} & 0.17 & 0.17 \\
\midrule
Average                & \textbf{0.89} & 0.49 & 0.51 \\
\bottomrule
\end{tabular}
\vspace{5mm}
\caption{Effect of data generation design on policy success rates. 
All variants use the same policy architecture and training budget; only the data generation procedure differs.}
\label{tab:g1_noise_ablation_psr}
\end{table}

\section{Real-World Experiment Details}
\label{app:real-world}

\textbf{Real world system.}
\label{app:real-world-hardware}
Each real-world policy is evaluated on 20 episodes.
We use a Luxonis OAK-D camera for image observations. For upper body joints, the control frequency is 25 Hz; for the lower body policy, inference is at 50 Hz. We use interpolation for the lowest-level position control API, which runs at 200 Hz.

\textbf{Scoring Criteria for Real World Experiments.}
\label{app:real-world-scoring}
Scoring criteria for \textbf{BoxToCart}: 0.5 for lifting box, 0.25 for walking to cart, 0.25 for placing on cart. Scoring criteria for \textbf{ThrowBottle}: 0.5 for lifting bottle, 0.5 for throwing bottle into bin. We award 0.5 for walking to the shelf and 0.5 for lifting the canister.

\end{document}